%% file: main.tex
\documentclass[runningheads]{llncs}

\usepackage{eccv}

\usepackage{eccvabbrv}

\usepackage{graphicx}
\usepackage{booktabs}
\usepackage{algorithm}
\usepackage{algpseudocode}

\usepackage[accsupp]{axessibility}  

\usepackage{hyperref}

\usepackage{orcidlink}

\newcommand{\cc}[1]{} 
\usepackage{booktabs}
\usepackage{multirow}
\usepackage{xcolor}
\usepackage{amssymb}
\definecolor{blueMPL}{HTML}{1f77b4}
\definecolor{orangeMPL}{HTML}{ff7f0e}
\definecolor{greenMPL}{HTML}{2ca02c}
\definecolor{redMPL}{HTML}{d62728}

\begin{document}

\title{Speed-Variation Distillation for Generative Modeling} 

\titlerunning{Speed-Variation Distillation for Generative Modeling}

\author{Sukit Seripanitkarn \and
Phonphrm Thawatdamrongkit \and \\
Supasorn Suwajanakorn$^*$ }

\authorrunning{S.~Seripanitkarn et al.}

\institute{VISTEC, Thailand \quad $^*$Corresponding author}

\maketitle

\vspace{-1.9em} 

\begin{figure}
    \centering
    \setlength{\tabcolsep}{4pt}
    \begin{tabular}{cc}
        \includegraphics[width=0.4\textwidth]{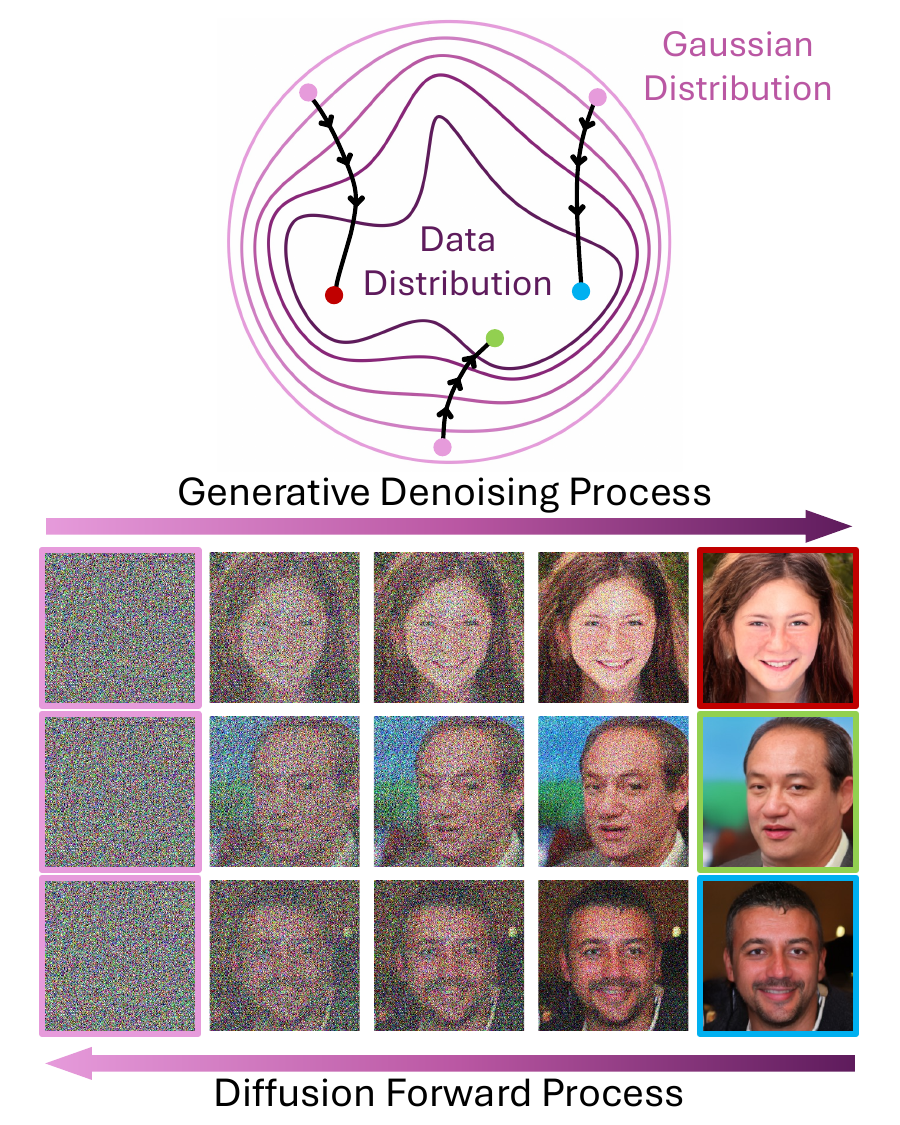}
        &
        \includegraphics[width=0.4\textwidth]{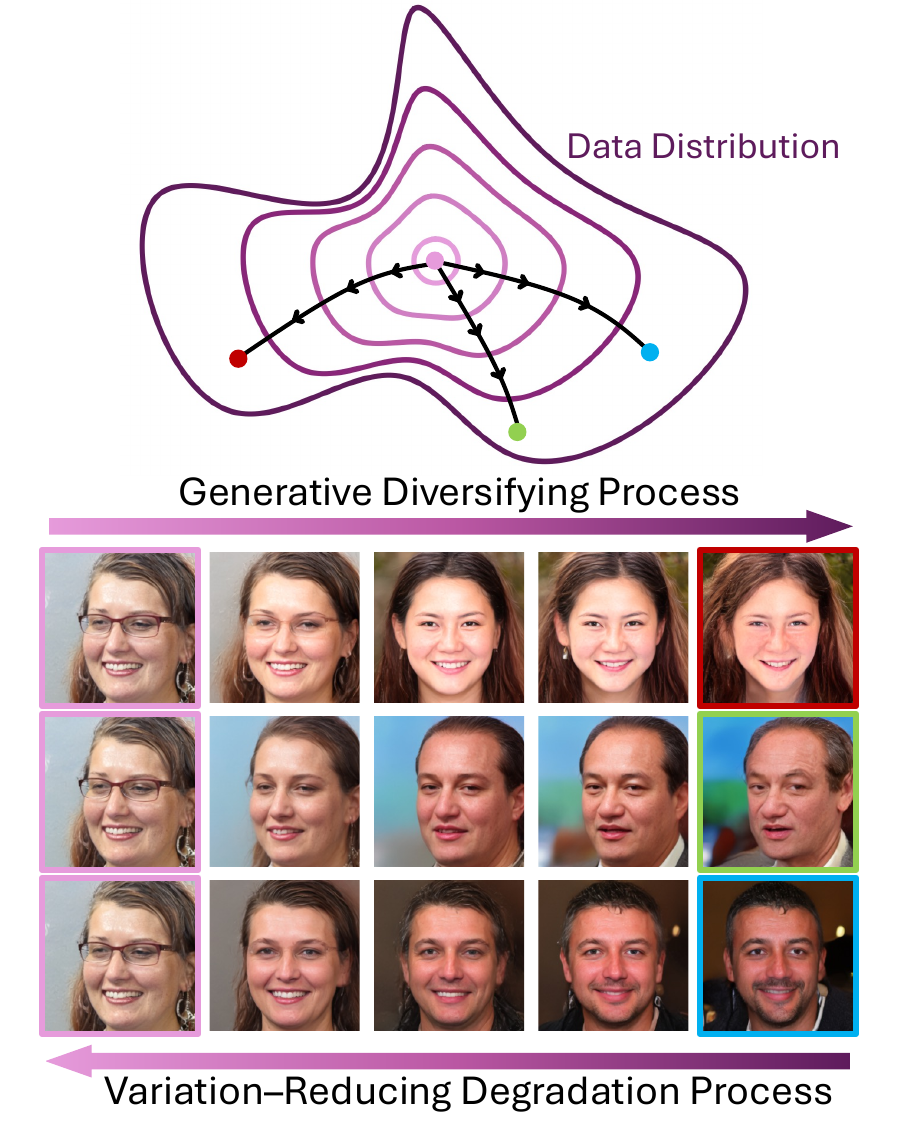}
        \\
        \textbf{(a) Diffusion Process}
        &
        \textbf{(b) Our Degradation Process}
    \end{tabular}
    \vspace{-0.7em}
    \caption{
    (a) Standard diffusion generates images by denoising Gaussian noise toward the data manifold, producing corrupted intermediate states for most of the trajectory.
    (b) Our method instead starts from a shared prototype already on the data manifold and reverses our variation-reducing degradation process, moving the point within the data distribution to gradually diversify it, enabling a controllable speed–variation trade-off.
    }

    
    
    \label{fig:teaser}
\end{figure}

\vspace{-3em}
\begin{abstract}
  While diffusion models excel at image generation, their denoising trajectories pass through corrupted intermediate states far from the image manifold. We ask whether generation can instead proceed through recognizable images throughout the entire trajectory. We introduce a generative framework built around a variation-reducing degradation process that contracts diverse images toward a shared prototype while preserving recognizable intermediate states. Because such manifold traversal lacks a simple analytical form, we construct the trajectories using generative-prior-based image morphing. We then distill the reverse process into a diversification operator that transforms the prototype into increasingly varied samples. To resolve the one-to-many ambiguity of the reverse process, we associate each trajectory with its initial noise latent and apply perturbations to improve robustness against accumulated errors. At inference, every step yields a recognizable image, allowing early termination and enabling a controllable speed--variation trade-off. 
  Please visit our website: \url{https://speed-variance-modeling.github.io}
  \vspace{-0.5em}
\end{abstract}

\input{sec/1_intro}

\input{sec/2_related_work}
\input{sec/3_background}
\input{sec/4_method}
\input{sec/5_experiment}
\input{sec/6_discussion}

\input{sec/7_conclusion}


%
%
\bibliographystyle{splncs04}
\bibliography{main}
\end{document}

%% file: sec/1_intro.tex
\section{Introduction}
\label{sec:intro}

Diffusion~\cite{ho2020denoising, song2020denoising} and flow-matching~\cite{lipman2022flow} models have emerged as the dominant generative paradigm due to their stable training dynamics and high sample quality. They generate data by progressively transforming a simple prior distribution into the target data distribution through a sequence of denoising steps. 
However, because the intermediate states along this trajectory are corrupted by noise, they cannot be used as natural images during generation. This raises a question: \emph{Can we design a generative process whose entire trajectory consists of valid, clean images from the target distribution?}

Several studies, including Cold Diffusion~\cite{bansal2023cold}, Soft Diffusion~\cite{daras2022soft}, Scale-Space Diffusion~\cite{mukhopadhyay2026scalespacediffusion}, and physics-inspired diffusion models~\cite{rissanen2022generative,gruszczynski2025beyond}, have explored degradation processes beyond Gaussian noise to improve generation quality, efficiency, and interpretability. Yet, these methods still fail to produce recognizable intermediate images as they progressively corrupt the input and move it away from the data manifold. As a result, the generated trajectories remain visually uninterpretable.



Our goal is to design a degradation process whose intermediate states remain on the data manifold. Unlike prior approaches that corrupt images with noise~\cite{ho2020denoising,song2020denoising,lipman2022flow}, blur~\cite{hoogeboom2022blurring,rissanen2022generative}, or downsampling~\cite{abu2024udpm,mukhopadhyay2026scalespacediffusion}, we gradually contract image variation toward a single prototype image while preserving visual realism throughout. By learning to reverse this trajectory, our model yields a generative process that can be terminated early at any step to produce a fully valid image---introducing a novel trade-off that exchanges sample diversity for faster generation.

We present a first realization of this principle through a distillation framework that extracts variation-reducing trajectories from pretrained generative models.
Specifically, we construct a dataset of trajectories via generative-prior-based image morphing, progressively transforming teacher-generated images along the data manifold toward a single, shared prototype.
However, learning to reverse these trajectories is an ill-posed one-to-many mapping from the single prototype to diverse targets, causing standard regression to collapse into blurry average predictions. Drawing inspiration from consistency distillation~\cite{song2023consistencymodels, luo2023latent,yin2024onestep} and rectified flow distillation~\cite{liu2023instaflow}, we resolve this multi-modality by conditioning the model on the init noise latent of the teacher-generated target image. This anchors each trajectory and converts the multi-modal generative task into a deterministic, one-to-one mapping. Finally, to mitigate error accumulation during iterative generation, we inject small perturbations into the model input during both training and inference.

Our main contributions are twofold:
First, we introduce a novel variation-reducing degradation that contracts image variation toward a shared prototype while preserving visual realism along the data manifold. Second, we propose a distillation framework to learn its reverse process, establishing a novel trade-off between speed and sample diversity.

%% file: sec/2_related_work.tex
\section{Related Work}

\subsection{Diffusion Models Beyond Gaussian Noise}
Diffusion models~\cite{ho2020denoising, song2020denoising} and flow models~\cite{lipman2022flow} have emerged as the leading paradigms for generative modeling, achieving state-of-the-art image synthesis quality and outperforming GAN-based methods~\cite{dhariwal2021diffusion}. Their success stems from progressively transforming samples drawn from a simple prior, typically a Gaussian distribution, into natural images. Recent work investigates whether the addition of noise is necessary to bridge the gap between a simple reference distribution and the target data distribution. Cold Diffusion~\cite{bansal2023cold} and Soft Diffusion~\cite{daras2022soft} show that a wide range of degradations can be paired with a learned restoration process. Subsequent work has explored blur-based trajectories in particular. Inverse Heat Dissipation models~\cite{rissanen2022generative} use blurring to encourage coarse-to-fine generation, Blurring Diffusion~\cite{hoogeboom2022blurring} connects blur to non-isotropic Gaussian diffusion, and Warm Diffusion~\cite{hsueh2025warm} studies how to balance blur and noise while preserving proximity to the data manifold. Beyond Blur~\cite{gruszczynski2025beyond} proposes an advection-diffusion corruption process that improves both image diversity and generation quality. More recently, Scale Space Diffusion~\cite{mukhopadhyay2026scalespacediffusion} leverages scale-space theory to progressively downsample images during the forward process and restore their original resolution during the reverse process, improving generation efficiency. However, these methods are primarily designed to improve sample quality and still rely on corruption processes that progressively transform images into visually unrecognizable intermediate states. In this work, we introduce a degradation process that progressively reduces image variation toward a single prototype while preserving recognizable intermediate images.

\subsection{Image Morphing}
Our variation-reducing degradation process uses image morphing to transform each image toward a shared prototype while preserving realistic intermediate states. Classical morphing methods rely on correspondence-based warping and blending, often requiring manual annotations and producing ghosting artifacts~\cite{beier2023morph,bhatt2011morph,darabi2012morph,liao2014morph,shechtman2010morph}. Deep generative models with continuous semantic latent spaces, such as variational autoencoders~\cite{kingma2013VAE}, StyleGAN~\cite{karras2019stylegan,karras2020stylegan2}, and Diffusion Autoencoders~\cite{preechakul2022diffae}, enable image morphing through latent interpolation that produces plausible intermediate states, but are typically limited to their training domains. Recent methods instead leverage large-scale text-to-image diffusion priors. Wang and Golland~\cite{wang2023interpolating} interpolate inverted noises and textual-inversion embeddings with pose guidance. IMPUS~\cite{yang2024impus} additionally introduces pair-specific LoRA adaptation and perceptually uniform sampling. DiffMorpher~\cite{zhang2024diffmorpher} fits and interpolates endpoint-specific LoRAs without textual inversion.
In this work, we adopt DiffMorpher for its efficiency in our many-to-one setting, where the prototype LoRA is fitted once and reused across all trajectories. Note that our degradation framework is agnostic to the specific morphing technique.

%% file: sec/3_background.tex
\section{Background}
Diffusion models progressively add noise to an image during the forward process and learn to reverse this process by denoising the noisy image back to a clean image. Latent Diffusion Models ~\cite{rombach2022high} operate in the latent space of a Variational Autoencoder (VAE) ~\cite{kingma2013VAE} to enable efficient high-resolution image generation. This is achieved by first encoding the input image $x$ into a latent representation $z=z^{0}$ using the VAE encoder $\mathcal{E}$. The forward diffusion process then perturbs $z^{0}$ to a latent variable at diffusion timestep $\tau$, which can be expressed as:
\begin{equation}
\label{eq:bg_forward} z^{\tau} = \sqrt{\alpha_{\tau}}\, z^{0} + \sqrt{1-\alpha_{\tau}}\, 
\epsilon, \qquad \epsilon \sim \mathcal{N}(0,I), 
\end{equation}
where $\{\alpha_{\tau}\}_{\tau=0}^{T}$ denotes the signal schedule over $T$ diffusion timesteps.

During the backward process, the diffusion models predict noise $\hat{\epsilon}^{\tau}$ added in the input $z^{\tau}$, however, this predicted noise can be turn into predicted clean $z^{0}$ denoted as $\hat{z}^{0}$ using the following equation:

\begin{equation}
\label{eq:bg_cleanpred}
    \hat{z}
    =
    \hat{z}^{0}
    =
    \frac{
        z^{\tau}
        -
        \sqrt{1-\alpha_{\tau}}\,
        \hat{\epsilon}^{\tau}
    }{
        \sqrt{\alpha_{\tau}}
    }.
\end{equation}

%% file: sec/4_method.tex
\section{Method}
\label{sec:method}
\begin{figure}[t]
    \centering
    \captionsetup{skip=5pt}
\includegraphics[width=1.0\linewidth]{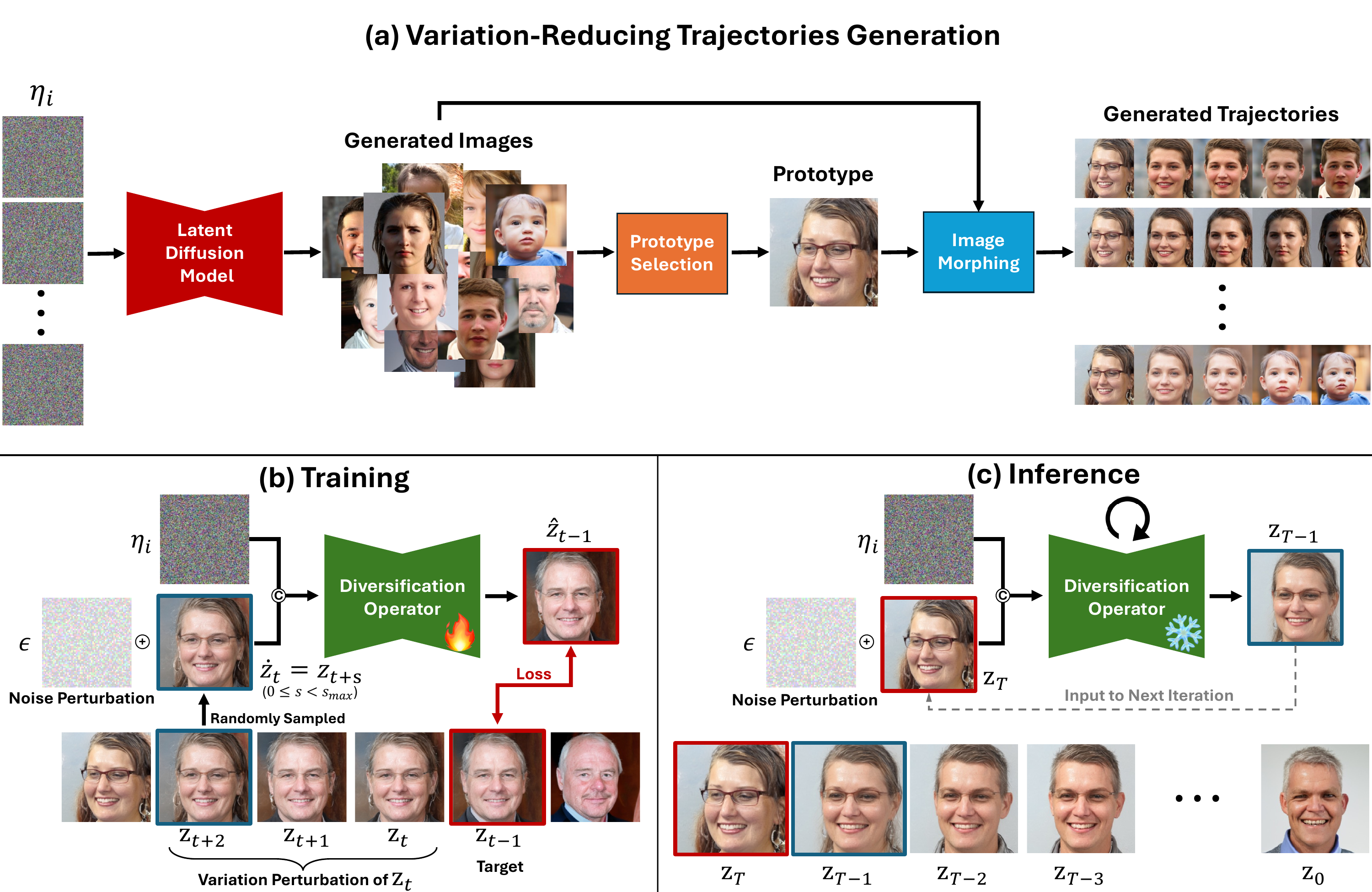}
\caption{\textbf{Overview.} (a) Generate images with a pretrained LDM, select a prototype, and construct variation-reducing trajectories. (b) We train the diversification operator using noise and variation perturbations to improve robustness during iterative generation, conditioned on the associated noise latent $\eta$. (c) The trained model generates clean, natural trajectories with a controllable speed-variation trade-off.}
    \label{fig:method}
    \vspace{-1em}
\end{figure}

We propose a distillation framework that learns a generative process with a controllable trade-off between sampling speed and output variation. Our framework consists of two main components. First, we define a variation-reducing degradation process that maps diverse images toward a shared prototype through realistic intermediate states, materialized as a curated trajectory dataset for efficient training. Second, we distill its reverse process into a diversification operator, employing noise conditioning and input perturbations to improve robustness during iterative generation. An overview of our method is shown in Fig.~\ref{fig:method}.

\subsection{Variation-Reducing Degradation Process}
We define a variation-reducing degradation process that progressively contracts diverse images toward a shared prototype through realistic intermediate states. Since this transformation traverse the image manifold rather than following an explicit corruption process, such as Gaussian noising, we realize it through image morphing with pretrained generative models to keep the trajectories close to the data manifold. Specifically, we first generate an image set and select a representative image as the shared prototype. We then connect each image to this prototype through a realistic morphing trajectory, producing trajectory data for training the reverse diversification process.

\subsubsection{Generating images and selecting the prototype.} We use a Latent Diffusion Model (LDM)~\cite{rombach2022high} to generate a set of images $\mathcal{X} = \{x_i\}$ together with the initial noise latent $\eta_i$ associated with each image. 
To select a representative prototype while preserving realism, we avoid pixel-space averaging, which produces blurry artifacts. Instead, we adopt a simple approach: we encode all images using a pretrained CLIP image encoder~\cite{radford2021learning} and select the prototype $x^{\star}$ as the image with the highest average cosine similarity to all other images in $\mathcal{X}$.


\vspace{-1em}
\subsubsection{Constructing image trajectories.}
To obtain smooth trajectories between the prototype and each target image, we use DiffMorpher~\cite{zhang2024diffmorpher}, an image morphing method, to generate image morphing sequences. Specifically, for each target image $x \in \mathcal{X}$, DiffMorpher generates a sequence of $T+1$ images that gradually transforms the target image into the prototype image $x^{\star}$:
\[
\{x_{0}, x_{1}, x_{2}, \ldots, x_{T}\},
\]
where $x_{0} = x$ is the target image and $x_{T} = x^{\star}$ is the prototype image. We repeat this process for every image in $\mathcal{X}$, yielding a set of trajectories connecting the shared prototype to all images in the dataset. 
Since generating these morphing trajectories requires costly inference through a generative prior, we precompute and cache them as a trajectory dataset for efficient model training.


\subsection{Diversification Operator}
Our goal is to distill the reverse of our variation-reducing trajectories into a diversification operator. Given an image $x_t$ at reduced variation, this operator predicts the higher-variation state $x_{t-1}$. Starting from the shared prototype $x_T$, iterative application of this operator acts as a generative process that yields increasingly diverse samples over progressive inference steps.

\vspace{-1em}
\subsubsection{Model input.}
We train our diversification operator by fine-tuning a Latent Diffusion Model (LDM) on the curated trajectory dataset. The model is initialized from the same pretrained LDM used to generate the initial image set and its associated initial noise latents.
Training is performed in the LDM latent space $\mathcal{Z}$, where each trajectory image $x_t$ is encoded by the VAE as $z_t \in \mathcal{Z}$.

Because the reverse mapping is one-to-many, $z_{t}$ alone leaves the diversification direction ambiguous, causing the model to average over plausible targets and produce blurry outputs. We therefore modify the model input by concatenating the associated initial noise latent $\eta$ with the current image latent $z_{t}$ along the channel dimension $C$ at trajectory step t, resulting in the input $\mathbf{x} = (z_{t}, \eta)$, where $\mathbf{x} \in \mathbb{R}^{2C \times H \times W}$. 

However, directly training on this input can lead to error accumulation, as small prediction errors may propagate and compound over successive inference steps. We improve robustness using two complementary strategies: \emph{variation perturbation} and \emph{noise perturbation}~\cite{laskey2017dart,chen2024diffusionforcing}. During training, we perform variation perturbation by randomly replacing the input $z_{t}$ with a lower-variation state $z_{t+s}$ from its local neighborhood, where $0 \le s < s_{\max}$ and $s_{\max}$ denotes the maximum perturbation range. The perturbed input state thus becomes $\dot{z}_t = z_{t+s}$. Next, we apply noise perturbation to $\dot{z}_t$ via the forward diffusion process at a fixed global timestep $\tau$ (Eq.~\ref{eq:bg_forward}), yielding the noisy latent $\dot{z}_t^\tau$. The complete input to the model is therefore:
\[
\mathbf{x} = (\dot{z}_{t}^{\tau}, \eta).
\]


In addition to the latent input, we condition the denoising network on
the current variation step $t$. To preserve the timestep-conditioning
mechanism of the pretrained LDM, we offset the variation timestep by the
fixed diffusion perturbation timestep $\tau$: $\tilde{t} = t + \tau$, and
$\tilde{t}$ is the timestep provided to the denoising network. The model
therefore predicts
\begin{equation}
    \hat{\epsilon}_t
    =
    \epsilon_{\theta}
    \left(
        \mathbf{x},
        \tilde{t}
    \right).
\end{equation}
This conditioning allows the model to distinguish different stages of
the diversification trajectory while remaining close to the timestep
conditioning used by the pretrained LDM.

\subsubsection{Training objective.}
During training, as outlined in Algorithm~\ref{alg:training}, we first sample a trajectory step $t$, with $z_{t-1}$ as the prediction target, and construct the input $\mathbf{x}=\left(\dot{z}_t^\tau,\eta\right)$.
Since our model is initialized from a pretrained LDM that is trained to predict the added noise, we retain $\epsilon-$parameterization during fine-tuning. Specifically, we first predict the noise residual $\hat{\epsilon}_t$ and then convert it to the associated clean latent $\hat{z}_{t-1}$ using Eq.~\ref{eq:bg_cleanpred}. Therefore, our training objective is: 
\begin{equation}
\label{eq:method_obj}
\mathcal{L}
=
\mathbb{E}_{t,z_{t},z_{t-1},\eta}
\left[
\left\|
\hat{z}_{t-1}
-
z_{t-1}
\right\|_1
\right].
\end{equation}

\vspace{-1em}
\subsubsection{Inference.} The model starts from the prototype latent $z_T$ and iteratively applies the learned diversification operator. Similar to training, we perturb the latent using the same $\tau$ value employed during training. The conditioning latent $\eta$ controls the direction of diversification. Since $\eta$ is the initial noise used to generate the target image during dataset construction, we can easily sample a new conditioning latent at inference time from the same Gaussian prior,
\[
\eta \sim \mathcal{N}(0,I).
\]
The number of inference steps controls how far the output moves away from the prototype: fewer steps produce samples closer to the prototype, while more steps produce increasingly diverse outputs. Finally, the resulting latent is decoded back to image space using the LDM VAE decoder. The complete inference procedure is summarized in Algorithm~\ref{alg:inference}.

\begin{figure*}[t]
    \centering
    \begin{minipage}[t]{0.56\textwidth}
        \begin{algorithm}[H]
            \caption{Training}
            \label{alg:training}
            \begin{algorithmic}[1]
                \Require Trajectory dataset $\mathcal{D}$, model $\epsilon_\theta$, noise perturbation $\tau$, variation perturbation $s_{max}$

                \While{not converged}
                    \State Sample $(z_t, z_{t-1}, \eta, t) \sim \mathcal{D}$
                    \State Sample $s \sim \{0,\ldots,s_{max}-1\}$
                    \State Set $\dot{z}_t \gets z_{t+s}$
                    \State Perturb $\dot{z}_t$ to obtain $\dot{z}_t^\tau$ using Eq.~\ref{eq:bg_forward}
                    \State Set $\tilde{t} \gets t + \tau$
                    \State $\hat{\epsilon}_t \gets
                    \epsilon_\theta(
                    \operatorname{concat}(\dot{z}_t^\tau,\eta),
                    \tilde{t})$
                    \State Recover $\hat{z}_{t-1}$ using Eq.~\ref{eq:bg_cleanpred}
                    \State Compute $\mathcal{L}$ using Eq.~\ref{eq:method_obj}
                    \State Update $\theta$
                \EndWhile
            \end{algorithmic}
        \end{algorithm}
    \end{minipage}
    \hfill
    \begin{minipage}[t]{0.42\textwidth}
        \begin{algorithm}[H]
            \caption{Inference}
            \label{alg:inference}
            \begin{algorithmic}[1]
                \Require Model $\epsilon_\theta$, noise perturbation $\tau$

                \State Sample $\eta \sim \mathcal{N}(0,I)$
                \State Prototype latent $z_T$ from $\epsilon_\theta$
                \For{$t = T, T-1, \ldots, 1$}
                    \State Perturb $z_t$ to obtain $z_t^\tau$ Eq.~\ref{eq:bg_forward}
                    \State Set $\tilde{t} \gets t + \tau$
                    \State $\hat{\epsilon}_t \gets
                    \epsilon_\theta(
                    \operatorname{concat}(z_t^\tau,\eta),
                    \tilde{t})$
                    \State Recover $z_{t-1}$ using Eq.~\ref{eq:bg_cleanpred}
                \EndFor
                \State Decode $\{z_t\}_{t=0}^{T}$ using the VAE
                \State \Return $\{x_t\}_{t=0}^{T}$
            \end{algorithmic}
        \end{algorithm}
    \end{minipage}
    \vspace{-1em}
\end{figure*}

%% file: sec/5_experiment.tex
\section{Experiments}
\subsection{Experiment Setup}

\subsubsection{Implementation details.} We use unconditional Latent Diffusion Models~\cite{rombach2022high} from the official repository as our base models. We set the number of variation steps to $T=50$ and the number of variation perturbation to $s_{max}=5$ for all datasets. Within a 1,000-step diffusion schedule, we use $\tau=300$ for FFHQ, CelebA, and LSUN-Bedrooms, and $\tau=200$ for LSUN-Churches. We fine-tune the models on the generated trajectory datasets using a batch size of 32 and a learning rate of $1\times10^{-5}$ for 20{,}000 training steps, except for LSUN-Churches, which is trained for 50{,}000 steps.

\vspace{-1em}
\subsubsection{Metrics.}
For quantitative evaluation, we use four metrics: Fréchet Inception Distance (FID)~\cite{heusel2017gans} to measure image fidelity, Inception Score (IS)~\cite{salimans2016improved} to assess sample diversity, and Precision (P) and Recall (R)~\cite{kynkaanniemi2019improved} to evaluate the coverage of the generated distribution. To evaluate intermediate states, we use ImageReward~\cite{xu2023imagereward} to measure perceptual image quality and the Vendi Score~\cite{friedman2022vendi} to measure sample diversity.

\vspace{-1em}
\subsubsection{Datasets.} We conduct most experiments on FFHQ, and additionally evaluate our method on CelebA, LSUN-Churches, and LSUN-Bedrooms. For each dataset, we generate 10,000 images using an unconditional Latent Diffusion Model with DDIM sampling and retain the corresponding initial noisy latent $\eta$ as the conditioning input to our model. All experiments are performed at a resolution of $256 \times 256$. Following Sec.~\ref{sec:method}, we select a prototype image for each dataset and use DiffMorpher~\cite{zhang2024diffmorpher} to construct a 50-step ($T=50$) trajectory that progressively transforms each generated image into the prototype.

\subsection{Main Results}
\begin{figure*}[!t]
  \centering
  \small
  \newcommand{\stripwidth}{0.96\linewidth}
  \begin{tabular}{cc}
    \begin{minipage}[t]{0.485\textwidth}
      \centering
      \textbf{(a) FFHQ}\\[0pt]
      \begin{tabular}{@{}c@{\hspace{2pt}}c@{}}
        \multirow{3}{*}{\rotatebox[origin=c]{90}{\small $x_{t}$}} & \includegraphics[width=\stripwidth]{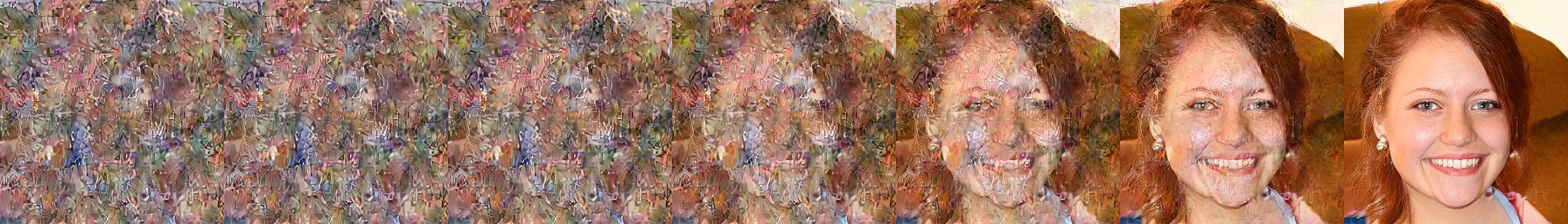}\\[-3pt]
                                                               & \includegraphics[width=\stripwidth]{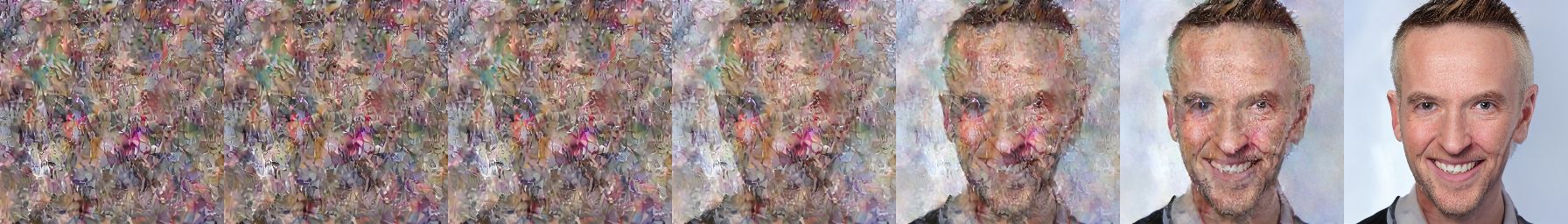}\\[-3pt]
                                                               & \includegraphics[width=\stripwidth]{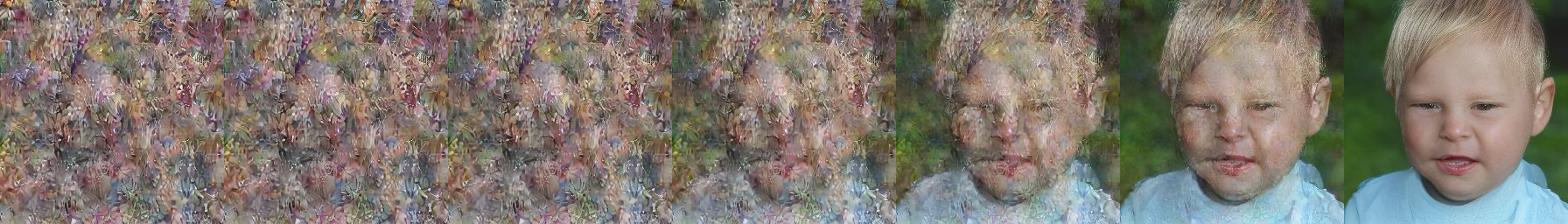}\\[3pt]
        \multirow{3}{*}{\rotatebox[origin=c]{90}{\small $\hat{x}_{0}$}} & \includegraphics[width=\stripwidth]{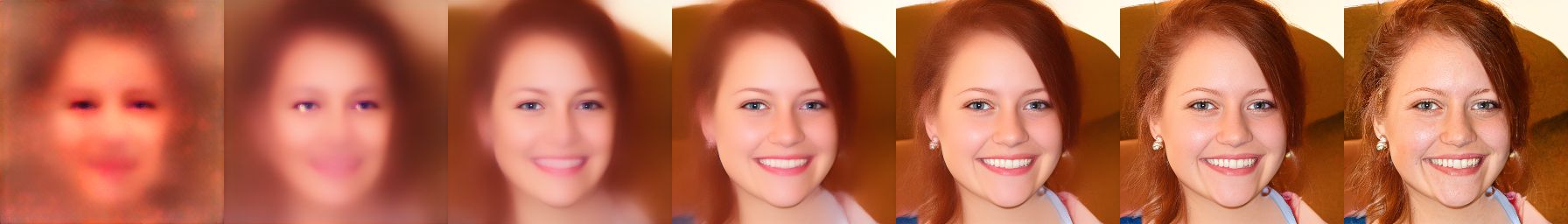}\\[-3pt]
                                                               & \includegraphics[width=\stripwidth]{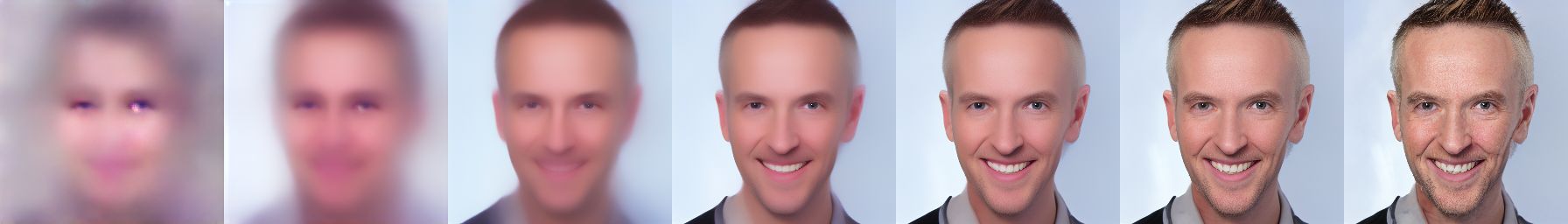}\\[-3pt]
                                                               & \includegraphics[width=\stripwidth]{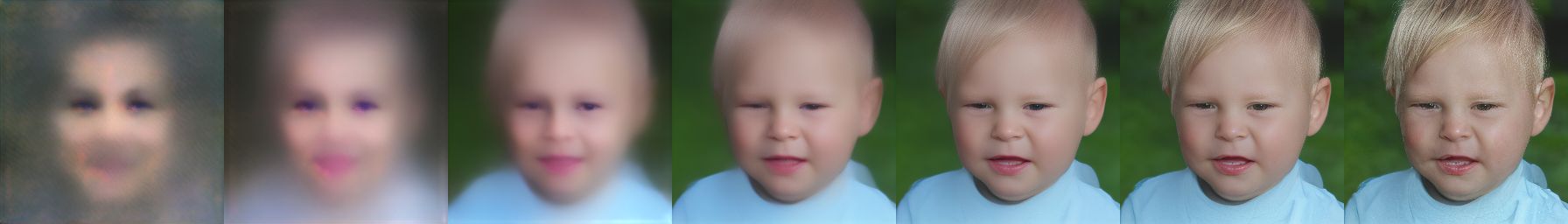}\\[3pt]
        \multirow{3}{*}{\rotatebox[origin=c]{90}{\small \textbf{Ours}}} & \includegraphics[width=\stripwidth]{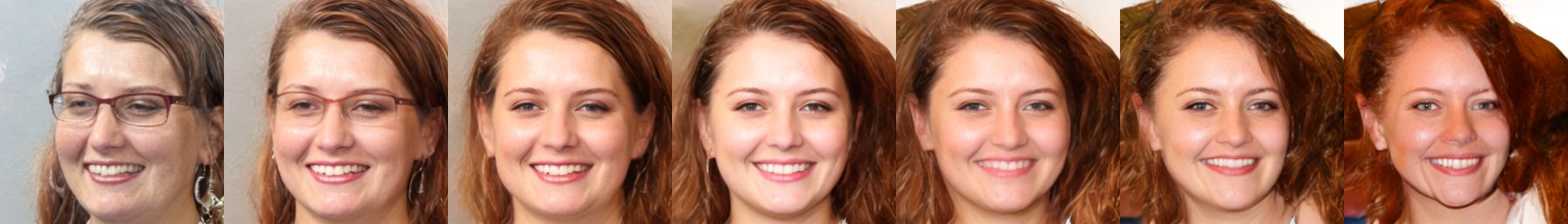}\\[-3pt]
                                                               & \includegraphics[width=\stripwidth]{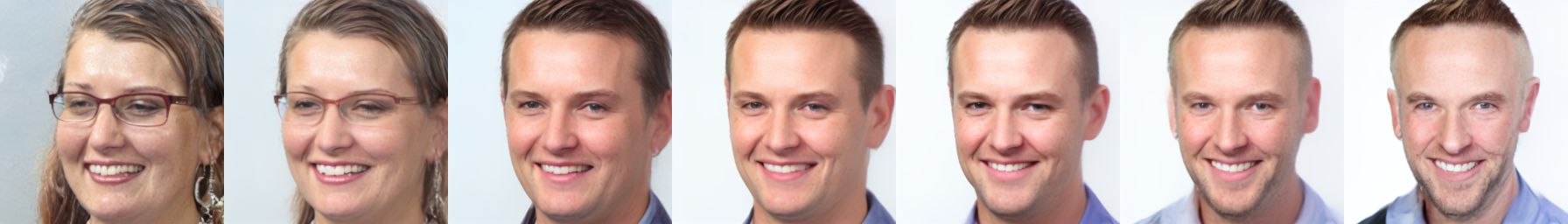}\\[-3pt]
                                                               & \includegraphics[width=\stripwidth]{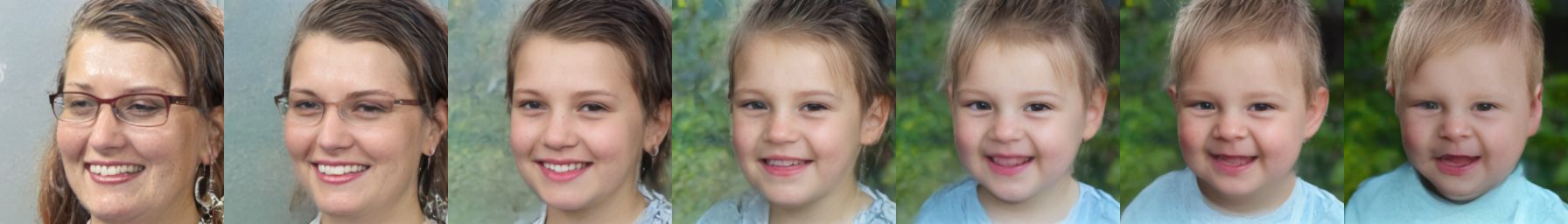}
      \end{tabular}
    \end{minipage}
    &
    \begin{minipage}[t]{0.485\textwidth}
      \centering
      \textbf{(b) CelebA}\\[0pt]
      \begin{tabular}{@{}c@{}}
        \includegraphics[width=\stripwidth]{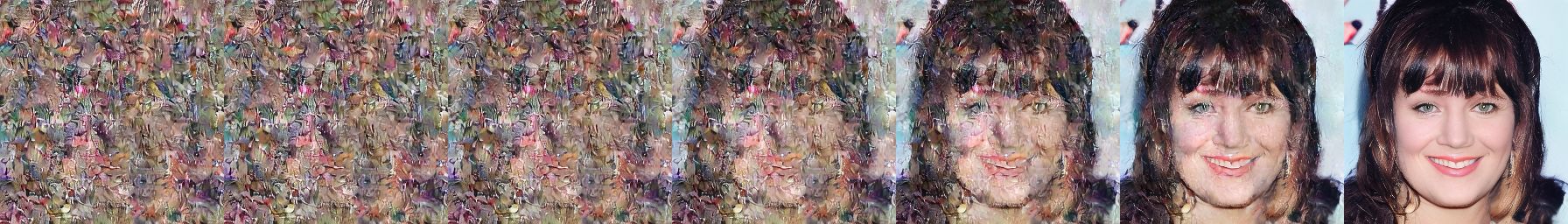}\\[-3pt]
        \includegraphics[width=\stripwidth]{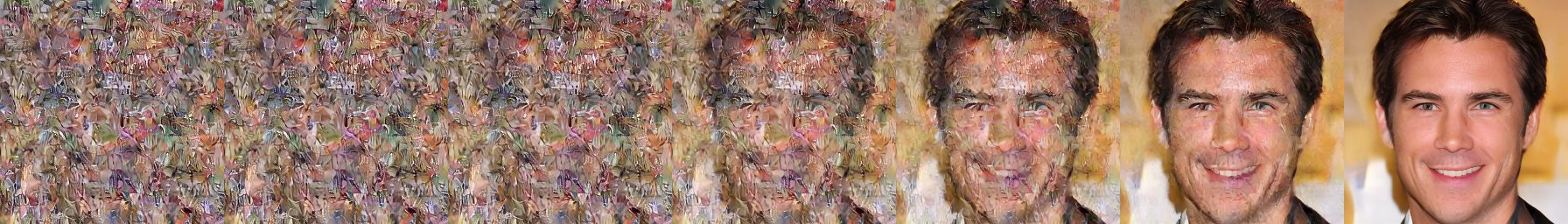}\\[-3pt]
        \includegraphics[width=\stripwidth]{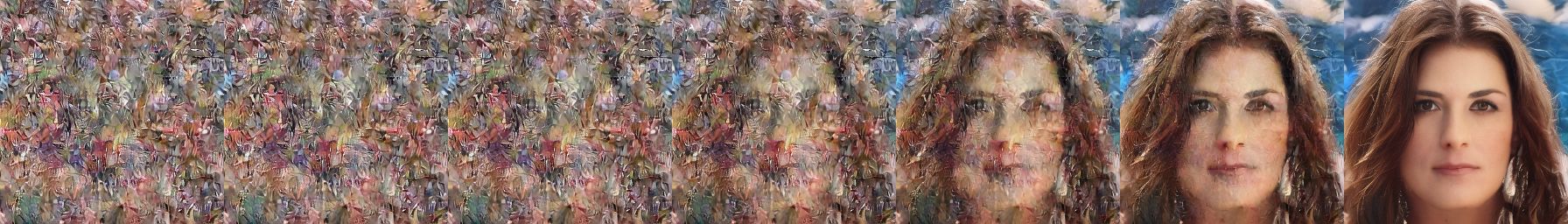}\\[3pt]
        \includegraphics[width=\stripwidth]{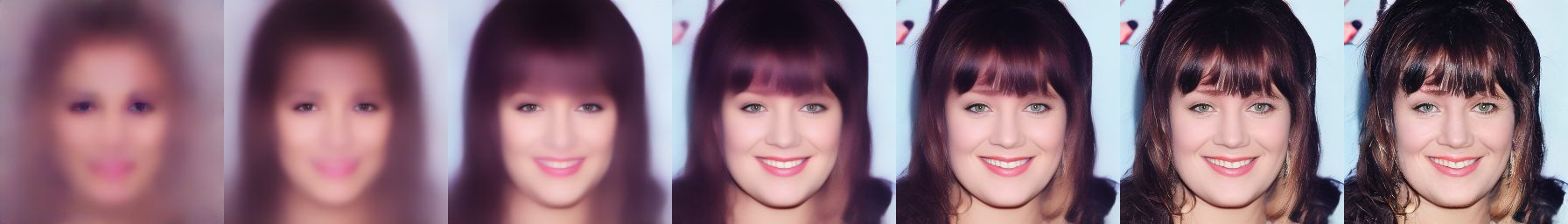}\\[-3pt]
        \includegraphics[width=\stripwidth]{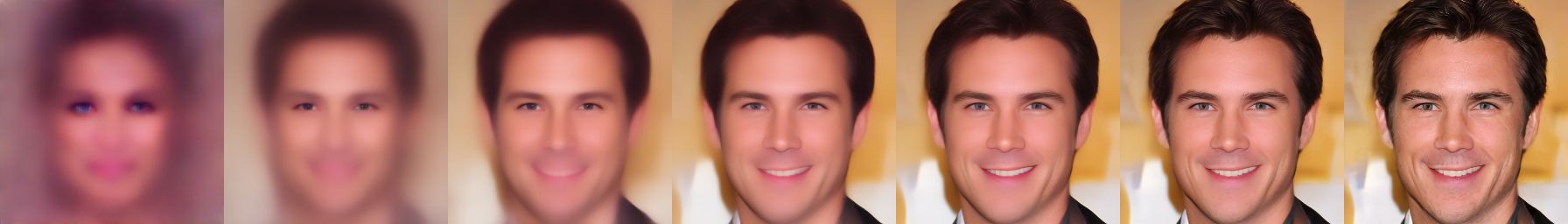}\\[-3pt]
        \includegraphics[width=\stripwidth]{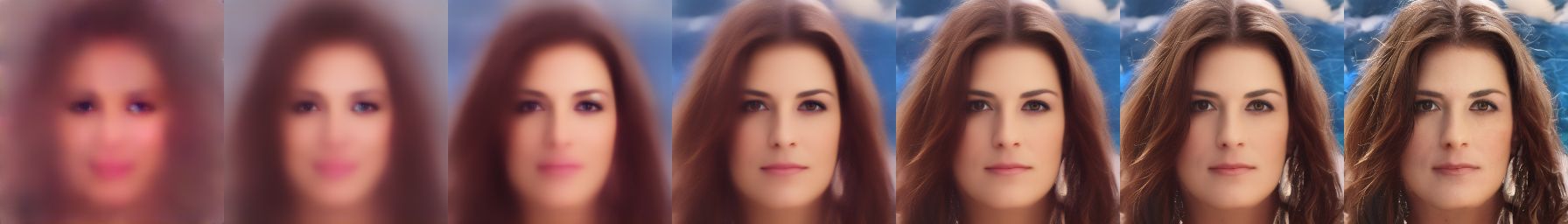}\\[3pt]
        \includegraphics[width=\stripwidth]{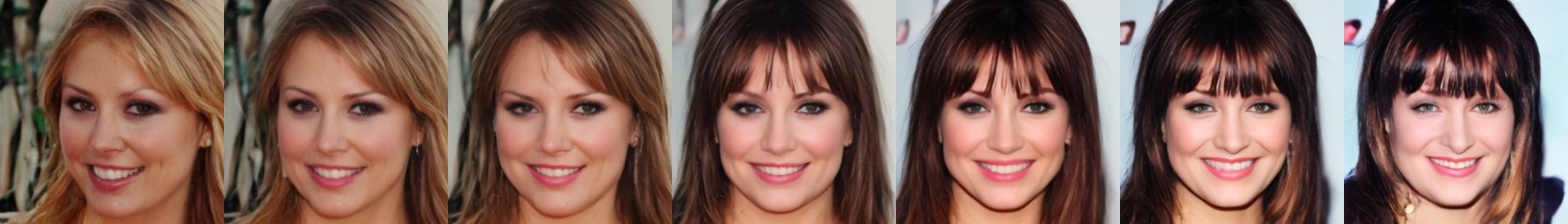}\\[-3pt]
        \includegraphics[width=\stripwidth]{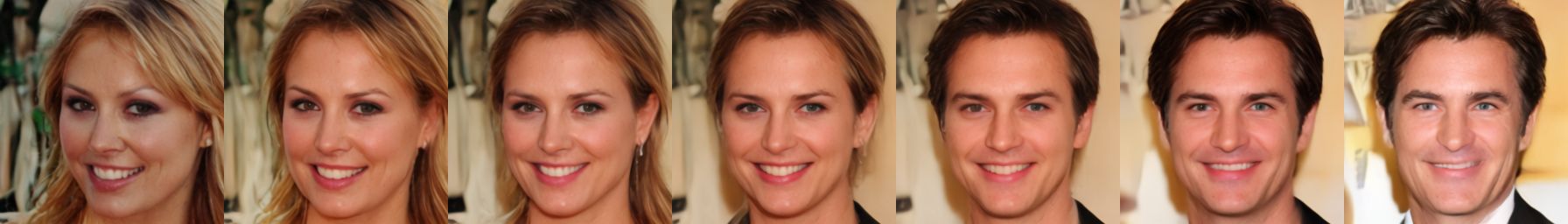}\\[-3pt]
        \includegraphics[width=\stripwidth]{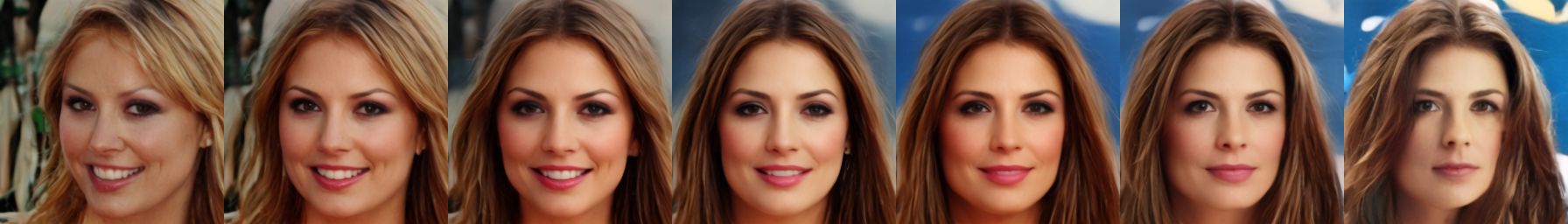}
      \end{tabular}
    \end{minipage}
    \\[12pt]
    \begin{minipage}[t]{0.485\textwidth}
      \centering
      \vspace*{1pt}
      \textbf{(c) Church}\\[0pt]
      \begin{tabular}{@{}c@{\hspace{2pt}}c@{}}
        \multirow{3}{*}{\rotatebox[origin=c]{90}{\small $x_{t}$}} & \includegraphics[width=\stripwidth]{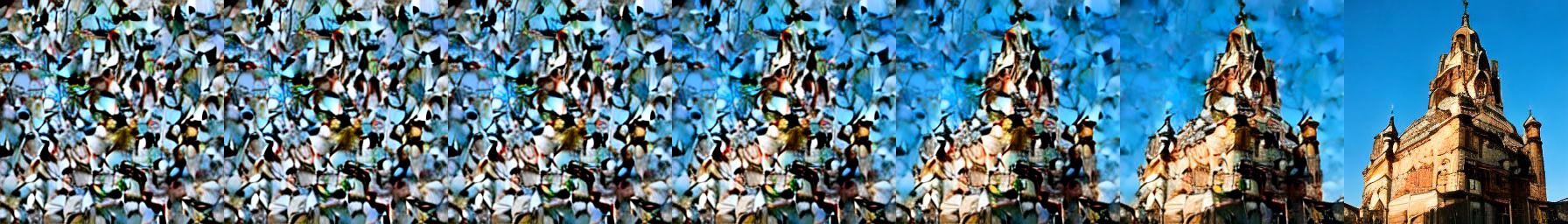}\\[-3pt]
                                                               & \includegraphics[width=\stripwidth]{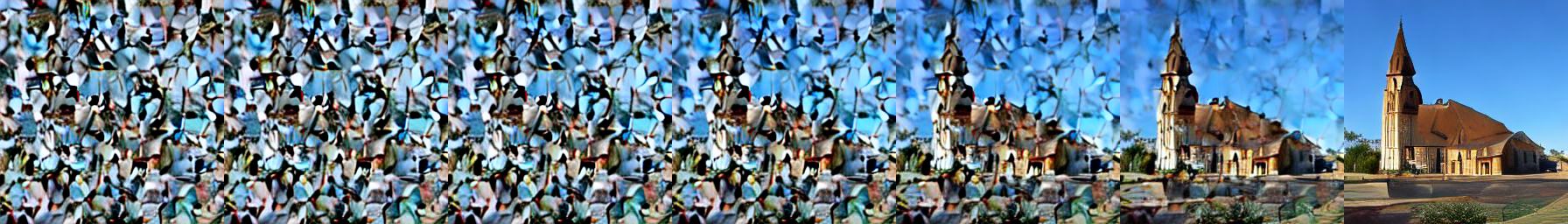}\\[-3pt]
                                                               & \includegraphics[width=\stripwidth]{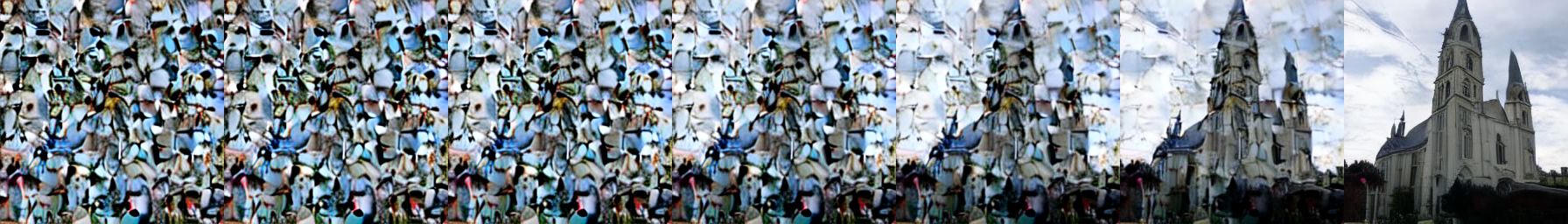}\\[3pt]
        \multirow{3}{*}{\rotatebox[origin=c]{90}{\small $\hat{x}_{0}$}} & \includegraphics[width=\stripwidth]{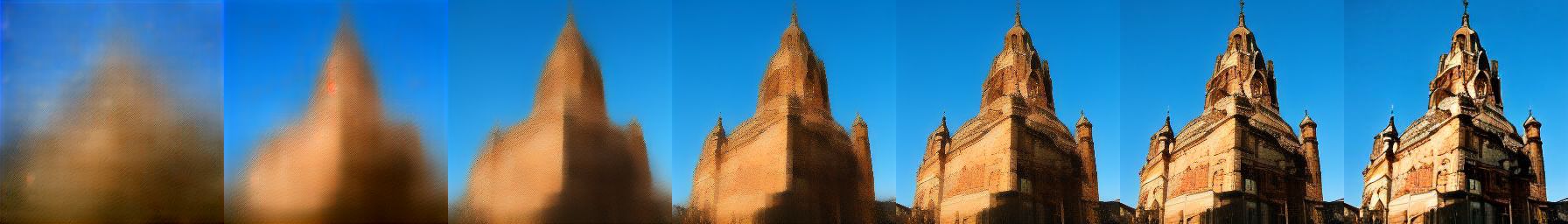}\\[-3pt]
                                                               & \includegraphics[width=\stripwidth]{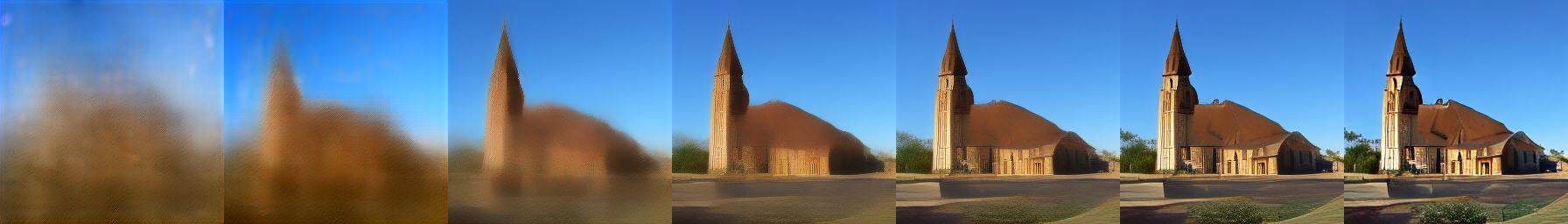}\\[-3pt]
                                                               & \includegraphics[width=\stripwidth]{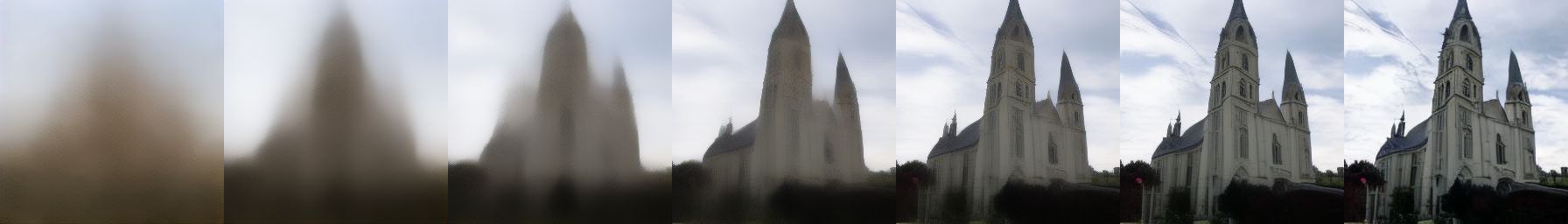}\\[3pt]
        \multirow{3}{*}{\rotatebox[origin=c]{90}{\small \textbf{Ours}}} & \includegraphics[width=\stripwidth]{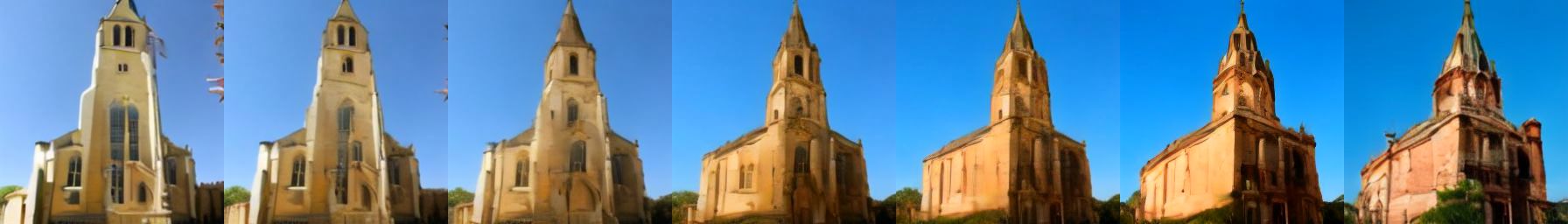}\\[-3pt]
                                                               & \includegraphics[width=\stripwidth]{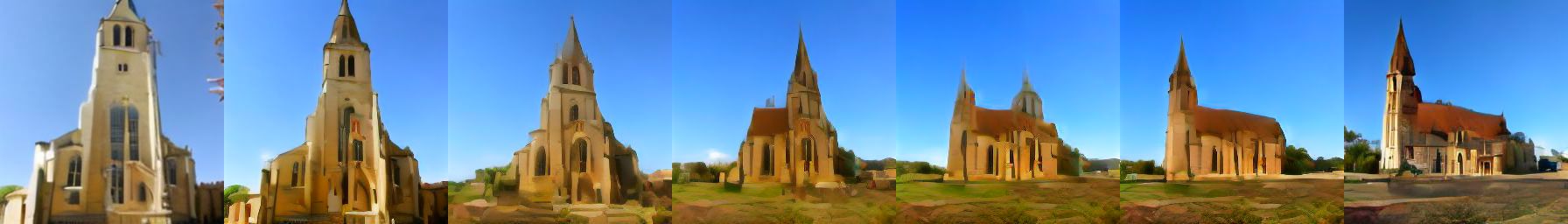}\\[-3pt]
                                                               & \includegraphics[width=\stripwidth]{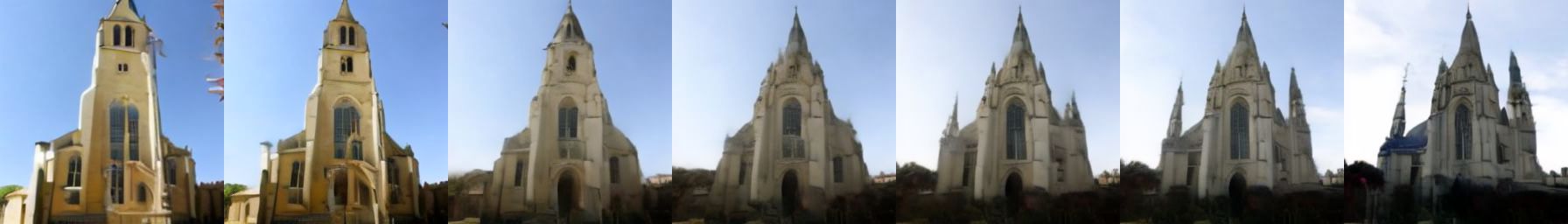}
      \end{tabular}
    \end{minipage}
    &
    \begin{minipage}[t]{0.485\textwidth}
      \centering
      \vspace*{1pt}
      \textbf{(d) Bedroom}\\[0pt]
      \begin{tabular}{@{}c@{}}
        \includegraphics[width=\stripwidth]{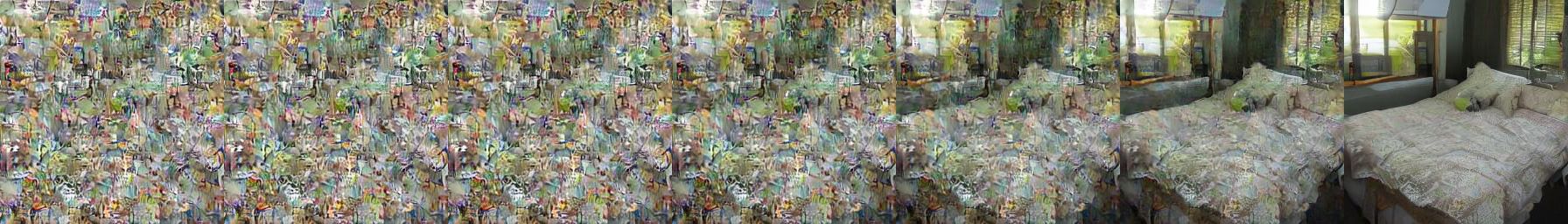}\\[-3pt]
        \includegraphics[width=\stripwidth]{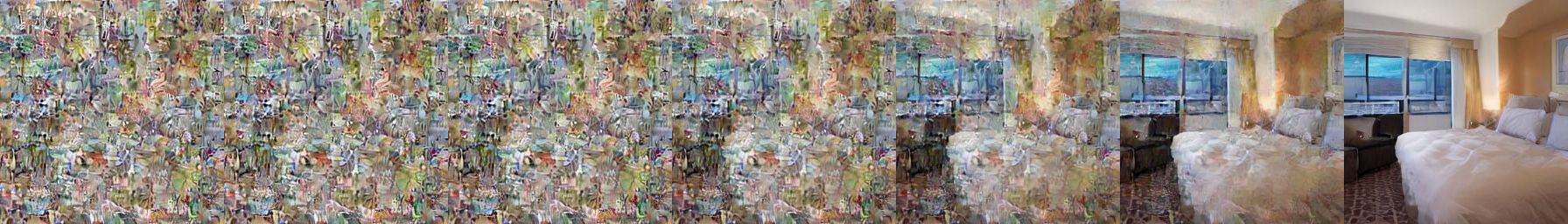}\\[-3pt]
        \includegraphics[width=\stripwidth]{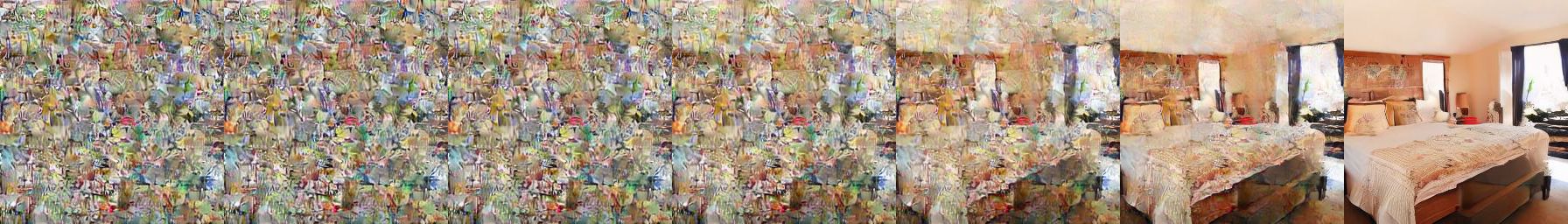}\\[3pt]
        \includegraphics[width=\stripwidth]{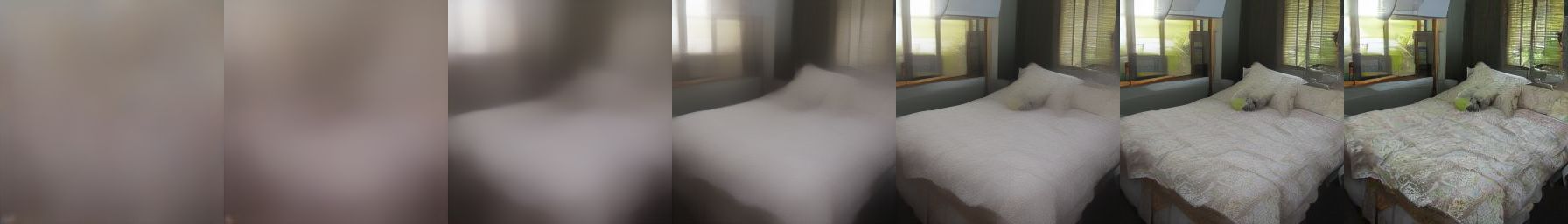}\\[-3pt]
        \includegraphics[width=\stripwidth]{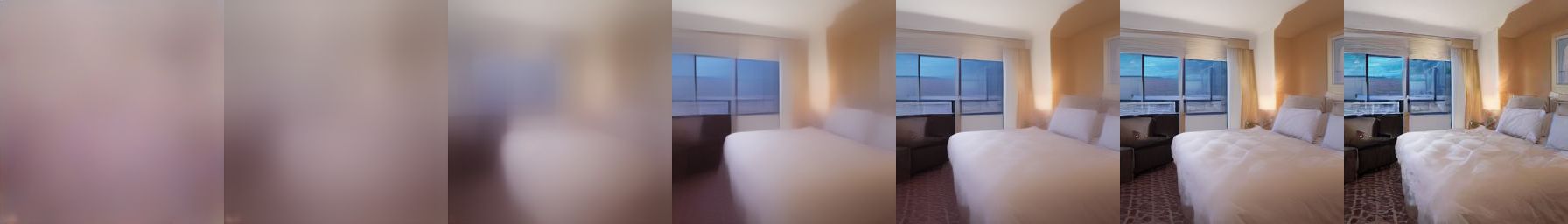}\\[-3pt]
        \includegraphics[width=\stripwidth]{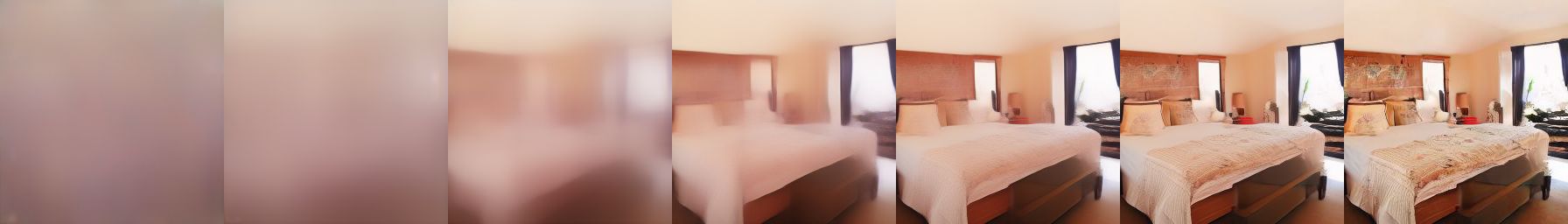}\\[3pt]
        \includegraphics[width=\stripwidth]{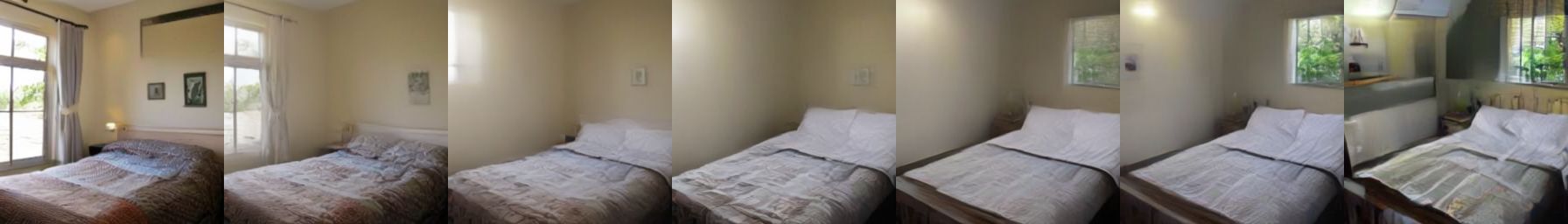}\\[-3pt]
        \includegraphics[width=\stripwidth]{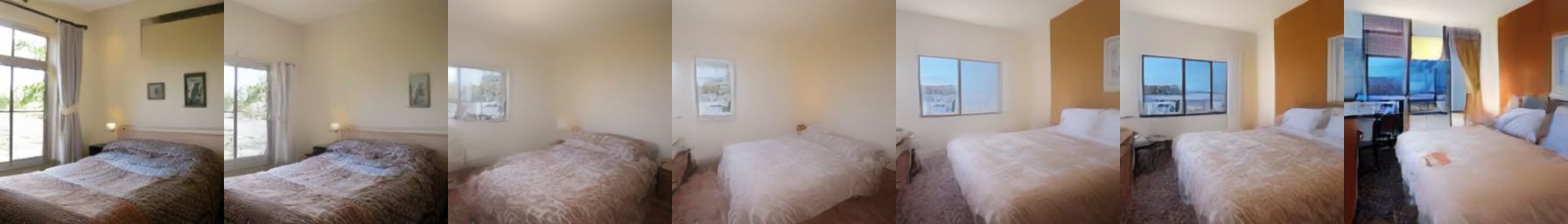}\\[-3pt]
        \includegraphics[width=\stripwidth]{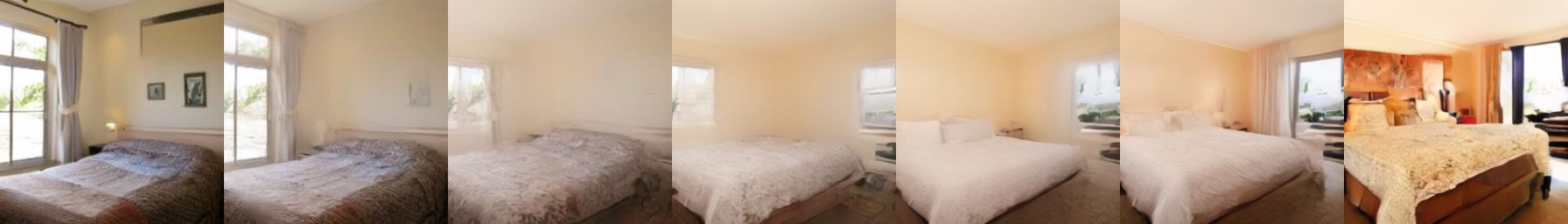}
      \end{tabular}
    \end{minipage}
  \end{tabular}
  \caption{Qualitative comparison of generated sequence strips across FFHQ, CelebA, Church, and Bedroom datasets. For each dataset grid, the top section shows $x_t$, the middle section shows $\hat{x}_0$, and the bottom section shows Ours.}
  \label{fig:main_results_strip}
\end{figure*}
We compare the trajectories produced by our diversification operator with those of a standard diffusion model (LDM). We visualize trajectories on FFHQ, CelebA, LSUN-Churches, and LSUN-Bedrooms, and evaluate 1{,}000 FFHQ samples per method using ImageReward and Vendi Score. For the diffusion baselines, we show the intermediate states $x_t$ and predicted clean image $\hat{x}_0$ at each denoising step, using a total of 50 deterministic DDIM steps.

Fig.~\ref{fig:main_results_strip} presents the comparison. Our method produces sharp, recognizable images from the earliest inference steps while progressively developing variations in appearance and high-level semantic attributes. For example, in the third FFHQ trajectory, the apparent age of the subject decreases gradually from an adult to a child. In contrast, the diffusion baseline determines this attribute early and primarily refines visual details afterward. More generally, the baseline remains blurry during early denoising stages and becomes visually coherent only toward the end of generation.

\begin{figure*}[t]
    \centering
    
    \includegraphics[width=0.95\textwidth]{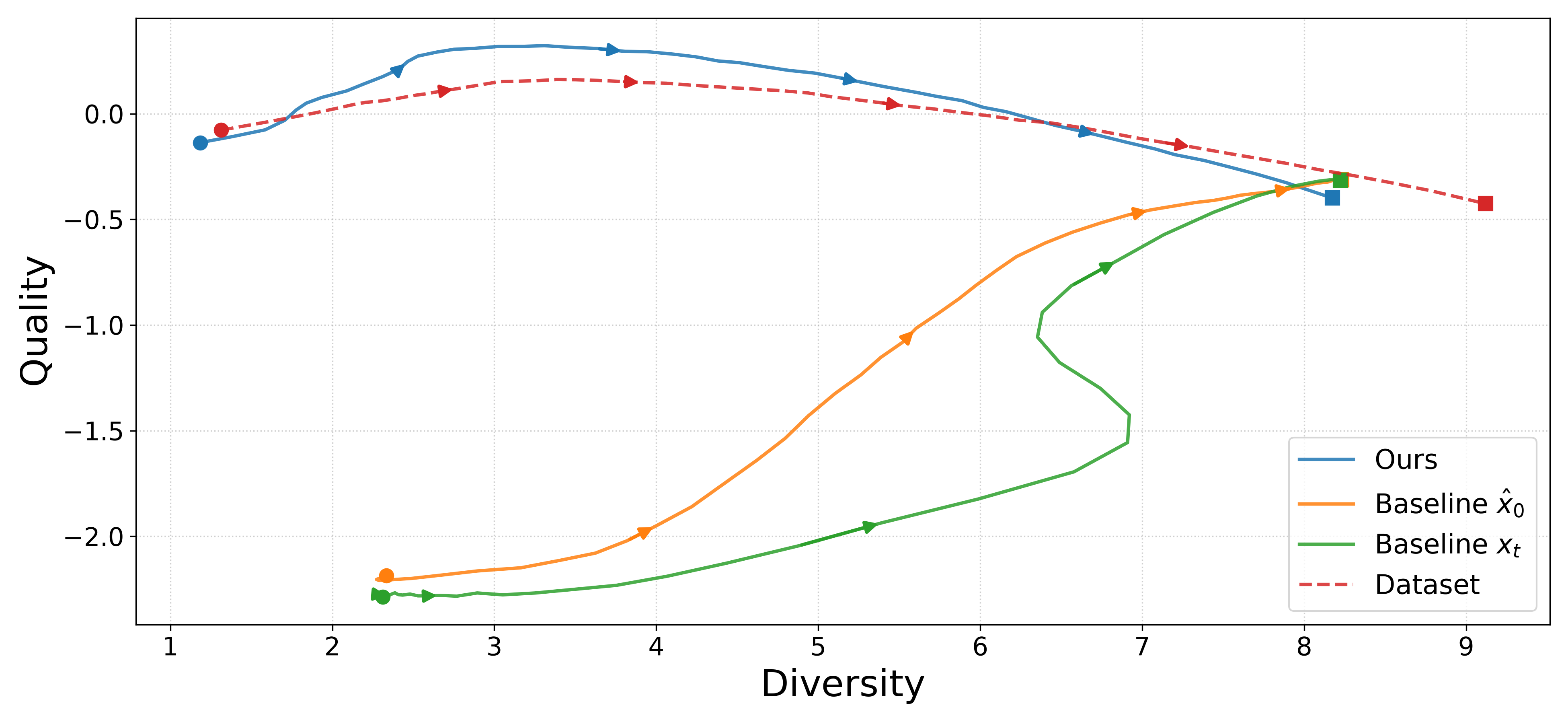} 
    \vspace{-10pt}
    \caption{
        \textbf{Image quality versus diversity across inference steps.}
        Image quality is measured using ImageReward, while diversity is measured using the Vendi Score.
        Arrows indicate the direction of generation, while circles (\textcolor{blueMPL}{$\bullet$}, \textcolor{redMPL}{$\bullet$}, \textcolor{orangeMPL}{$\bullet$}, \textcolor{greenMPL}{$\bullet$}) and squares mark (\textcolor{blueMPL}{$\blacksquare$}, \textcolor{redMPL}{$\blacksquare$}, \textcolor{orangeMPL}{$\blacksquare$}, \textcolor{greenMPL}{$\blacksquare$})
        the initial and final steps, respectively. We compare our trajectory with the
        diffusion baseline's intermediate states $x_t$ and predicted clean images
        $\hat{x}_0$. The dashed curve represents the variation-reducing trajectories in the curated training dataset. 
        Our method maintains high image quality as diversity progressively increases.
    }
    \label{fig:intermediate_states_plot}
\end{figure*}

Fig.~\ref{fig:intermediate_states_plot} illustrates the speed--variation trade-off by plotting image quality against sample diversity across inference steps. Our trajectory closely follows the variation-reducing trajectory in the curated training dataset, denoted by ``Dataset,'' while maintaining high image quality as diversity increases. In contrast, both the intermediate states $x_t$ and predicted clean images $\hat{x}_0$ of the diffusion baseline begin at lower quality (\textcolor{greenMPL}{$\bullet$}, \textcolor{orangeMPL}{$\bullet$}) and improve only during later denoising stages. These results show that our model produces high-quality samples early (\textcolor{blueMPL}{$\bullet$}) while progressively increasing diversity, enabling a controllable speed--variation trade-off.

\vspace{-1em}
\subsubsection{Importance of noise association $\eta$.}
\begin{figure*}[!t]
    \centering
    \setlength{\tabcolsep}{2pt}
    \renewcommand{\arraystretch}{1.0}
    \begin{tabular}{c c c c c c c}
        & \textbf{$t=T=50$} 
        & \textbf{$t=40$} 
        & \textbf{$t=30$} 
        & \textbf{$t=20$} 
        & \textbf{$t=10$} 
        & \textbf{$t=0$} \\
        
        \raisebox{0.8\height}{\rotatebox[origin=c]{90}{\text{Without $\eta$}}}
        & \includegraphics[width=0.15\textwidth]{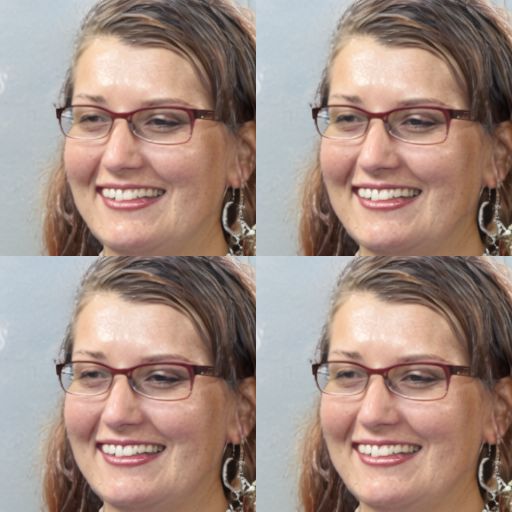}
        & \includegraphics[width=0.15\textwidth]{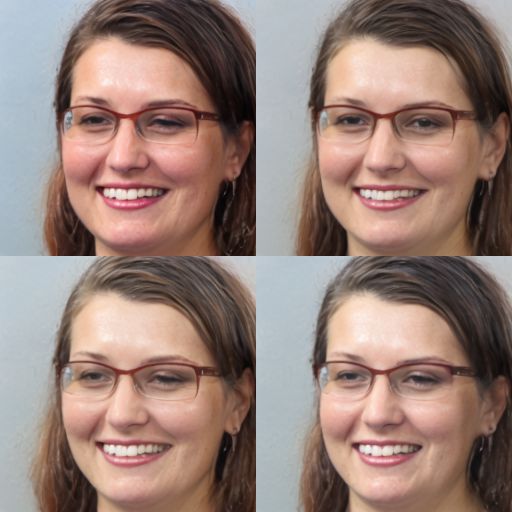}
        & \includegraphics[width=0.15\textwidth]{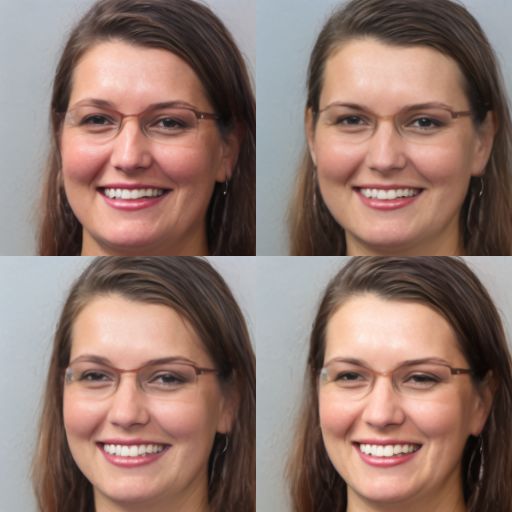}
        & \includegraphics[width=0.15\textwidth]{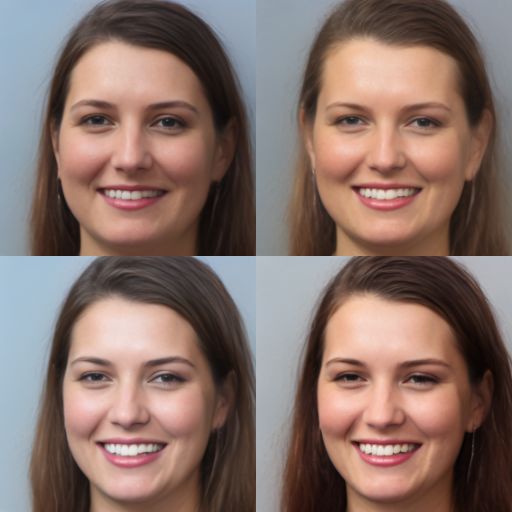}
        & \includegraphics[width=0.15\textwidth]{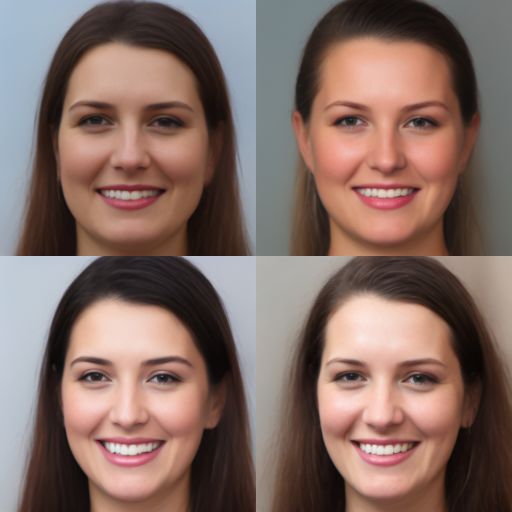}
        & \includegraphics[width=0.15\textwidth]{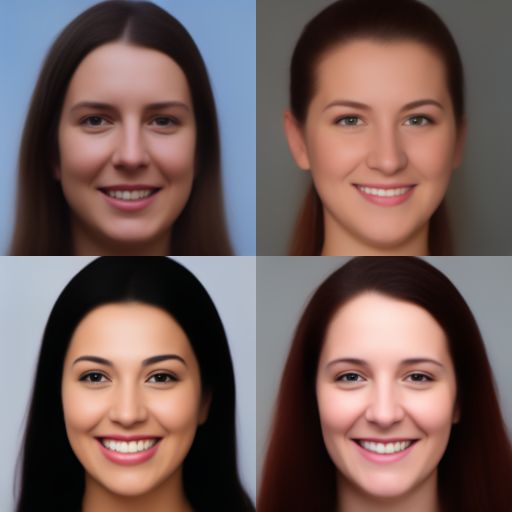} \\

        \raisebox{1.4\height}{\rotatebox[origin=c]{90}{\text{With $\eta$}}}
        & \includegraphics[width=0.15\textwidth]{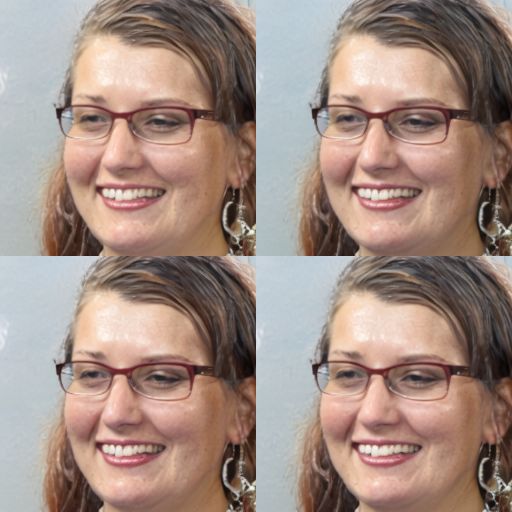}
        & \includegraphics[width=0.15\textwidth]{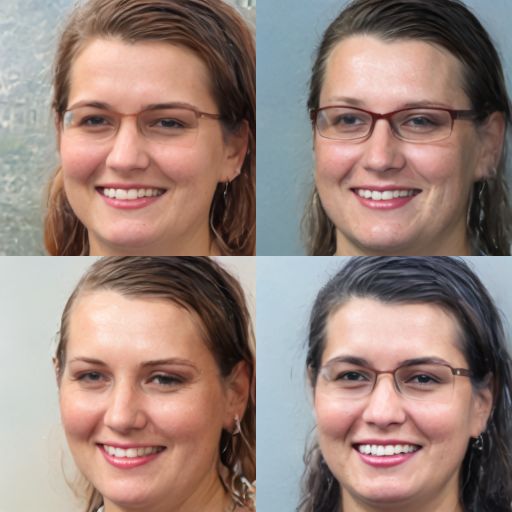}
        & \includegraphics[width=0.15\textwidth]{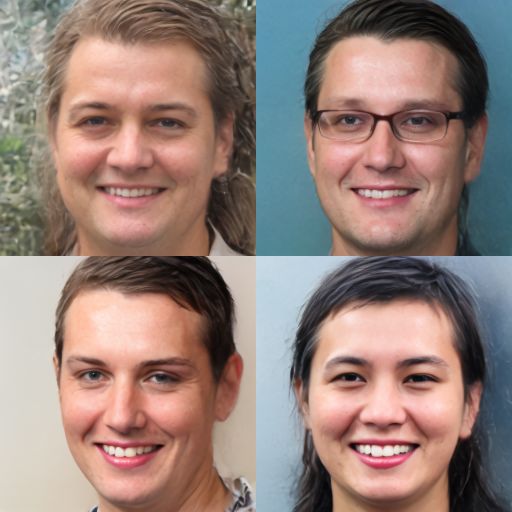}
        & \includegraphics[width=0.15\textwidth]{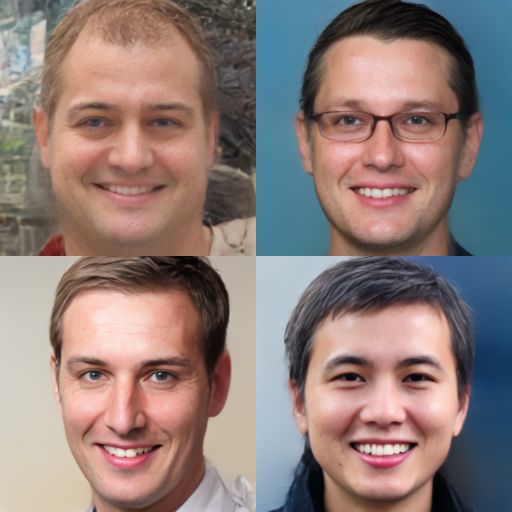}
        & \includegraphics[width=0.15\textwidth]{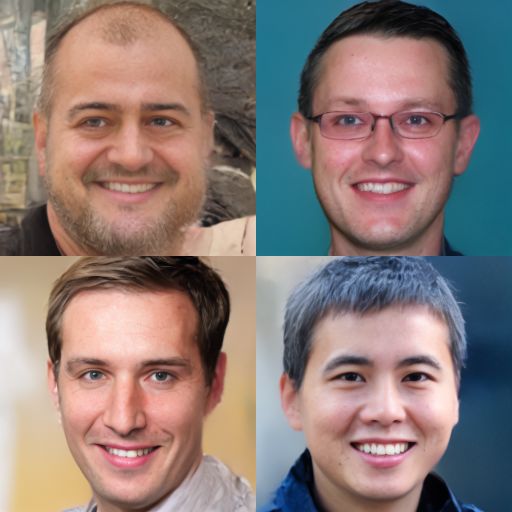}
        & \includegraphics[width=0.15\textwidth]{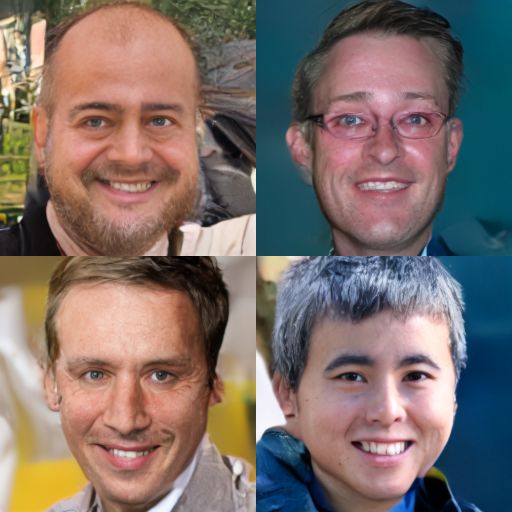} \\
    \end{tabular}
    \vspace{-10pt}
    \caption{
    \textbf{Comparison of our method with and without noise association $\eta$.} Noise association $\eta$ enables greater variation throughout the trajectory, whereas omitting it results in trajectories with limited variation and minimal changes.
    }
    \label{fig:associated_noise_ablation}
    \vspace{-1em}
\end{figure*}
We show the importance of noise association $\eta$ in our method by removing $\eta$ from the input to the model, leaving only the perturbed input image, to construct an unconditional version of our method. Fig.~\ref{fig:associated_noise_ablation} shows that without $\eta$, the generated samples exhibit lower variation throughout the trajectory and change much less than those with $\eta$. This shows that noise association $\eta$ acts as a guiding direction toward specific semantics, leading the model to generate correct attributes, such as changing the prototype from female to male, while without $\eta$, the model struggles to do so.

\vspace{-1em}
\subsubsection{Importance of variation perturbation $s_{max}$.}
\begin{figure*}[!t]
    \centering
    \setlength{\tabcolsep}{1pt}
    \renewcommand{\arraystretch}{0.8}
    \begin{tabular}{c cccccc}
        & \textbf{\small $t=T=50$} & \textbf{\small $t=40$} & \textbf{\small $t=30$} & \textbf{\small $t=20$} & \textbf{\small $t=10$} & \textbf{\small $t=0$} \\
        \raisebox{1.0\height}{\rotatebox[origin=c]{90}{\small $s_{max}=1$}}
        & \includegraphics[width=0.13\textwidth]{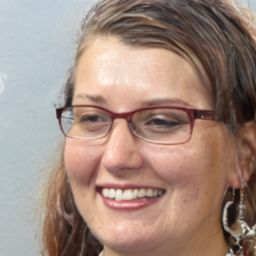}
        & \includegraphics[width=0.13\textwidth]{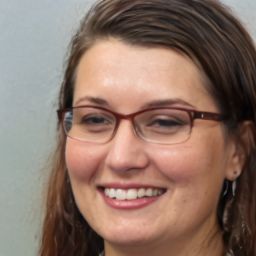}
        & \includegraphics[width=0.13\textwidth]{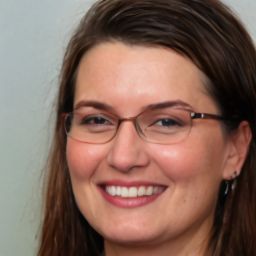}
        & \includegraphics[width=0.13\textwidth]{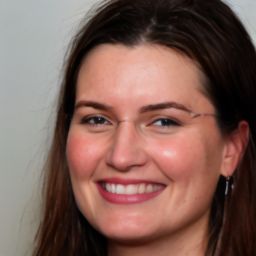}
        & \includegraphics[width=0.13\textwidth]{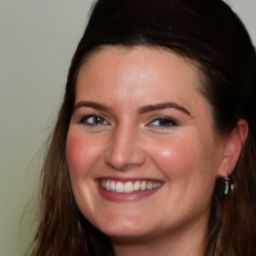}
        & \includegraphics[width=0.13\textwidth]{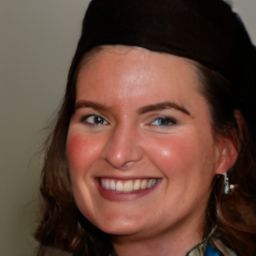} \\

        \raisebox{0.9\height}{\rotatebox[origin=c]{90}{\small $s_{max}=5$}}
        & \includegraphics[width=0.13\textwidth]{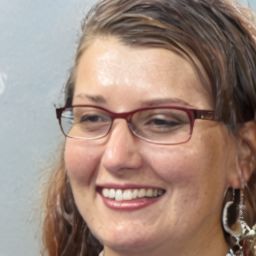}
        & \includegraphics[width=0.13\textwidth]{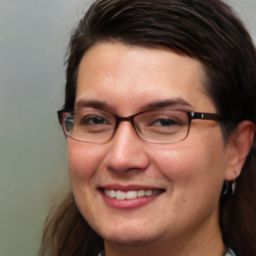}
        & \includegraphics[width=0.13\textwidth]{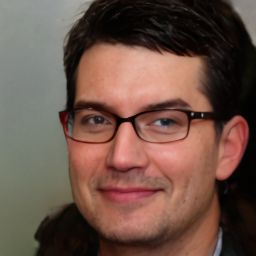}
        & \includegraphics[width=0.13\textwidth]{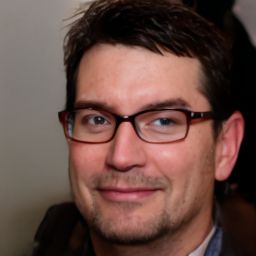}
        & \includegraphics[width=0.13\textwidth]{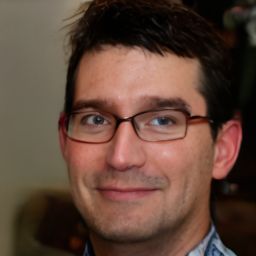}
        & \includegraphics[width=0.13\textwidth]{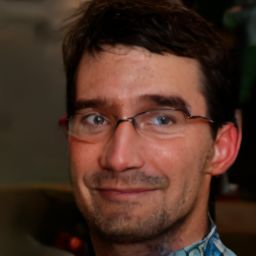} \\

        \raisebox{0.75\height}{\rotatebox[origin=c]{90}{\small\shortstack{Ground\\Truth}}}
        & \includegraphics[width=0.13\textwidth]{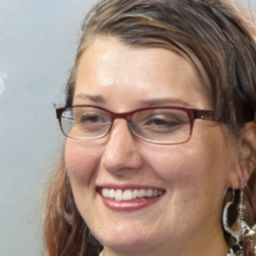}
        & \includegraphics[width=0.13\textwidth]{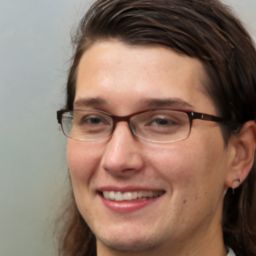}
        & \includegraphics[width=0.13\textwidth]{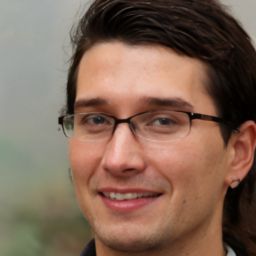}
        & \includegraphics[width=0.13\textwidth]{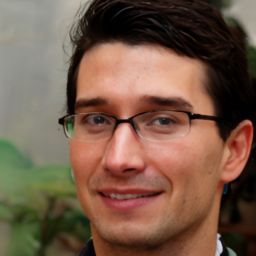}
        & \includegraphics[width=0.13\textwidth]{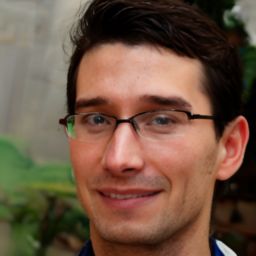}
        & \includegraphics[width=0.13\textwidth]{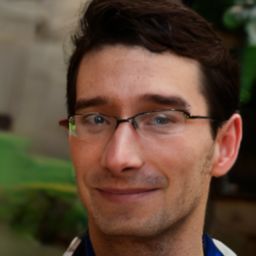} \\
    \end{tabular}
    \vspace{-2mm}
    \caption{\small \textbf{Comparison of our method with ($s_{max}=5$) and without ($s_{max}=1$) variation perturbation $s_{max}$.} Using variation perturbation produces trajectories that more faithfully match the curated ground-truth trajectories.}
    \label{fig:support_ablation}
    \vspace{-1em}
\end{figure*}
We analyze the effect of the maximum variation perturbation range $s_{max}$ in Table~\ref{tab:s_ablation}. The results show that using no variation perturbation ($s_{max}=1$) leads to a high FID score, indicating poorer generation quality. As shown in Fig.~\ref{fig:support_ablation}, without variation perturbation, 
the intermediate trajectory generation changes more slowly compared to the ones with support $s_{max}=5$ and ground truth, leading to poorer generation. We find that $s_{max}=5$ performs best in our experiments, achieving the lowest FID and highest IS, while larger values of $s_{max}$ progressively degrade FID performance.

\vspace{-1em}
\subsubsection{Importance of noise perturbation $\tau$.}
\begin{figure*}[t]
    \centering
    \setlength{\tabcolsep}{1pt}
    \renewcommand{\arraystretch}{0.8}
    \begin{tabular}{c cccccc}
        & \textbf{\small $t=T=50$} & \textbf{\small $t=40$} & \textbf{\small $t=30$} & \textbf{\small $t=20$} & \textbf{\small $t=10$} & \textbf{\small $t=0$} \\
        \raisebox{1.2\height}{\rotatebox[origin=c]{90}{\small $\tau=0$}}
        & \includegraphics[width=0.13\textwidth]{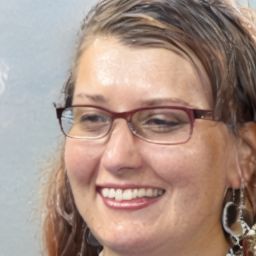}
        & \includegraphics[width=0.13\textwidth]{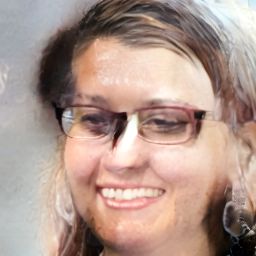}
        & \includegraphics[width=0.13\textwidth]{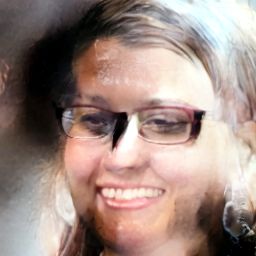}
        & \includegraphics[width=0.13\textwidth]{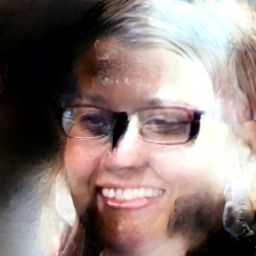}
        & \includegraphics[width=0.13\textwidth]{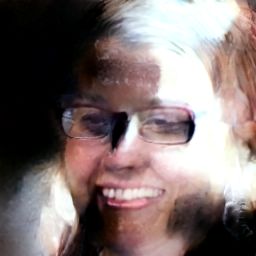}
        & \includegraphics[width=0.13\textwidth]{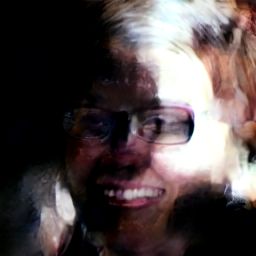} \\

        \raisebox{1.2\height}{\rotatebox[origin=c]{90}{\small $\tau=100$}}
        & \includegraphics[width=0.13\textwidth]{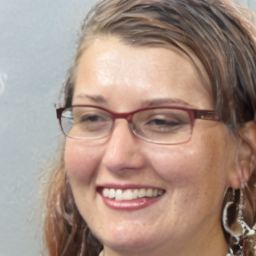}
        & \includegraphics[width=0.13\textwidth]{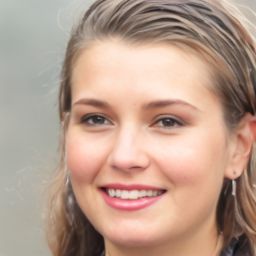}
        & \includegraphics[width=0.13\textwidth]{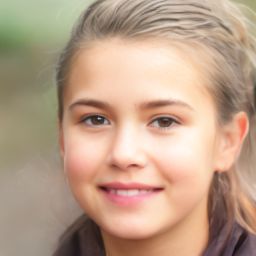}
        & \includegraphics[width=0.13\textwidth]{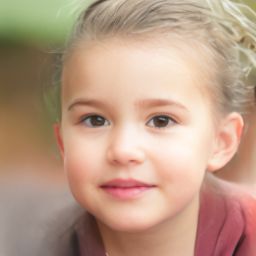}
        & \includegraphics[width=0.13\textwidth]{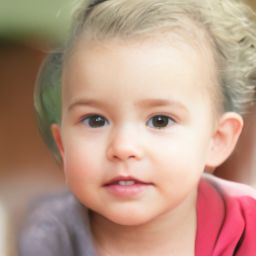}
        & \includegraphics[width=0.13\textwidth]{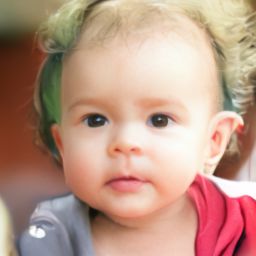} \\

        \raisebox{1.2\height}{\rotatebox[origin=c]{90}{\small $\tau=300$}}
        & \includegraphics[width=0.13\textwidth]{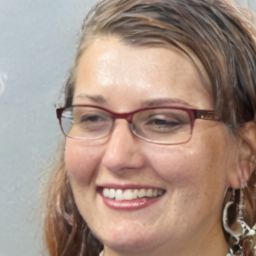}
        & \includegraphics[width=0.13\textwidth]{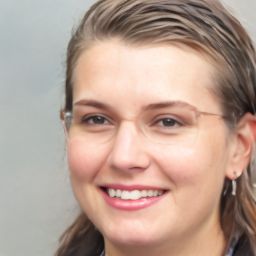}
        & \includegraphics[width=0.13\textwidth]{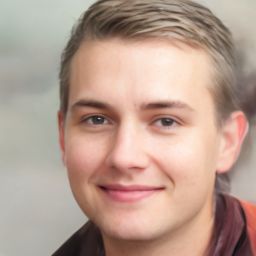}
        & \includegraphics[width=0.13\textwidth]{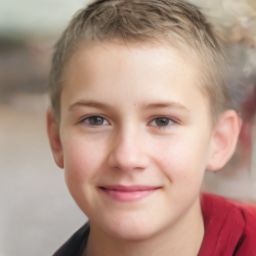}
        & \includegraphics[width=0.13\textwidth]{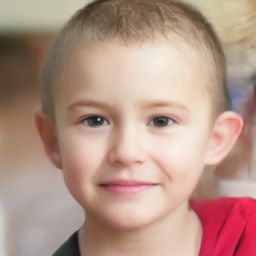}
        & \includegraphics[width=0.13\textwidth]{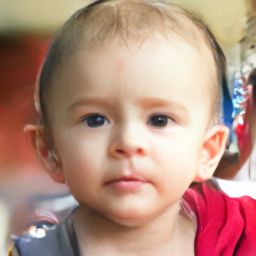} \\

        \raisebox{1.2\height}{\rotatebox[origin=c]{90}{\small $\tau=500$}}
        & \includegraphics[width=0.13\textwidth]{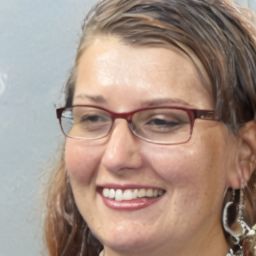}
        & \includegraphics[width=0.13\textwidth]{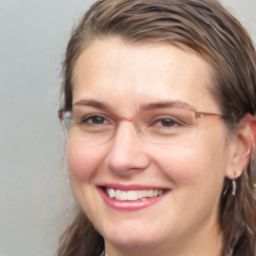}
        & \includegraphics[width=0.13\textwidth]{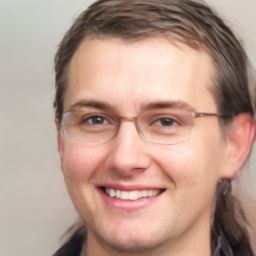}
        & \includegraphics[width=0.13\textwidth]{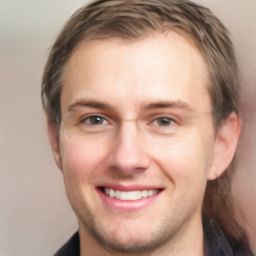}
        & \includegraphics[width=0.13\textwidth]{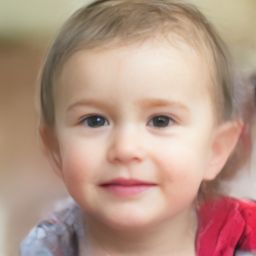}
        & \includegraphics[width=0.13\textwidth]{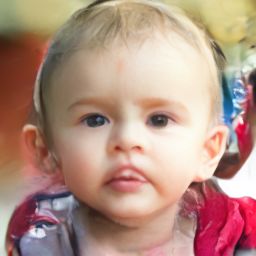} \\

        \raisebox{1.2\height}{\rotatebox[origin=c]{90}{\small $\tau=800$}}
        & \includegraphics[width=0.13\textwidth]{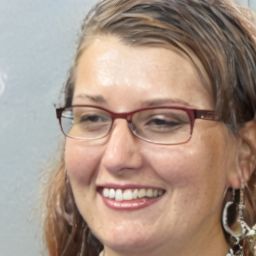}
        & \includegraphics[width=0.13\textwidth]{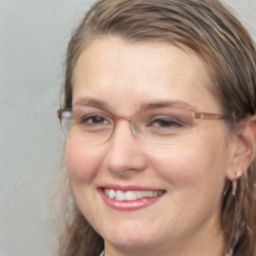}
        & \includegraphics[width=0.13\textwidth]{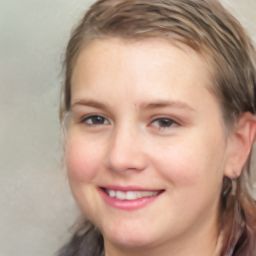}
        & \includegraphics[width=0.13\textwidth]{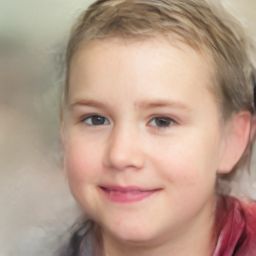}
        & \includegraphics[width=0.13\textwidth]{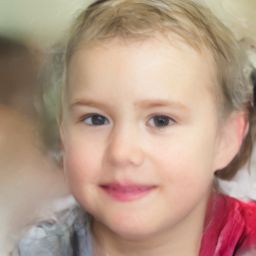}
        & \includegraphics[width=0.13\textwidth]{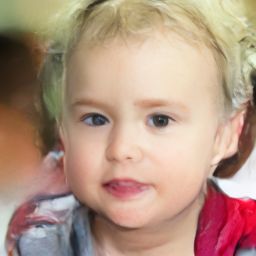} \\
    \end{tabular}
    \vspace{-2mm}
    \caption{\small \textbf{Effect of noise perturbation $\tau$ on image quality.} Both insufficient ($\tau=0$) and excessive ($\tau=500$ and $\tau=800$) noise perturbation degrade image quality, resulting in corrupted, distorted, or blurry images. In contrast, a moderate noise perturbation ($\tau=300$) provides the best balance and produces the highest-quality results with lowest FID as shown in Table~\ref{tab:nl_ablation}.}
    \label{fig:noise_timestep_ablation}
    \vspace{-1em}
\end{figure*}

We show the impact of different noise perturbation levels $\tau$ quantitatively in Table~\ref{tab:nl_ablation} and qualitatively in Fig.~\ref{fig:noise_timestep_ablation}. 
Without noise perturbation ($\tau=0$), the model struggles to learn the trajectory generation, leading to corrupted outputs. Conversely, excessively high noise levels, such as $\tau=500$ and $\tau=800$, introduce severe artifacts, distortion, and blurriness. We find that $\tau=300$ performs best, achieving the lowest FID score.

\begin{table}[t]
  \centering
  \begin{minipage}[t]{0.48\linewidth}
    \centering
    \caption{Quantitative results under different numbers of noise perturbation ($\tau$).}
    \label{tab:nl_ablation}
    \begin{tabular}{c c c c c}
      \toprule
      $\tau$ & FID $\downarrow$ & IS $\uparrow$ & P $\uparrow$ & R $\uparrow$ \\
      \midrule
      0   & 199.018 & 3.178 & 0.068 & 0.000 \\
      100 & 47.664  & 3.105 & 0.771 & 0.116 \\
      300 & \textbf{25.407} & \textbf{3.912} & 0.758 & 0.166 \\
      500 & 30.530  & 3.825 & 0.804 & 0.114 \\
      800 & 40.552  & 3.427 & \textbf{0.816} & 0.060 \\
      \bottomrule
    \end{tabular}
  \end{minipage}
  \hfill
  \begin{minipage}[t]{0.48\linewidth}
    \centering
    \caption{Quantitative results under different variation perturbation $s_{max}$.}
    \label{tab:s_ablation}
    \begin{tabular}{c c c c c}
      \toprule
      $s_{max}$ & FID $\downarrow$ & IS $\uparrow$ & P $\uparrow$ & R $\uparrow$ \\
      \midrule
      1  & 46.120 & 2.729 & \textbf{0.866} & 0.098 \\
      5  & \textbf{25.407} & \textbf{3.912} & 0.758 & \textbf{0.166} \\
      10 & 27.868 & 3.584 & 0.793 & 0.144 \\
      20 & 30.501 & 3.488 & 0.801 & 0.115 \\
      50 & 32.405 & 3.466 & 0.780 & 0.093 \\
      \bottomrule
    \end{tabular}
  \end{minipage}
\end{table}




%% file: sec/6_discussion.tex
\section{Limitations and Discussion}






\subsubsection{Data quality.}
\begin{figure*}[!t]
    \centering
    \includegraphics[width=0.6\linewidth]{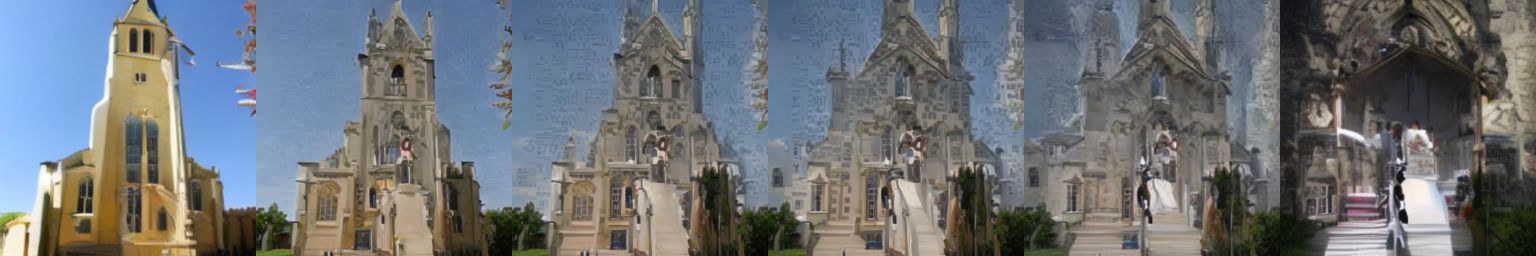} \\[1pt]
    \includegraphics[width=0.6\linewidth]{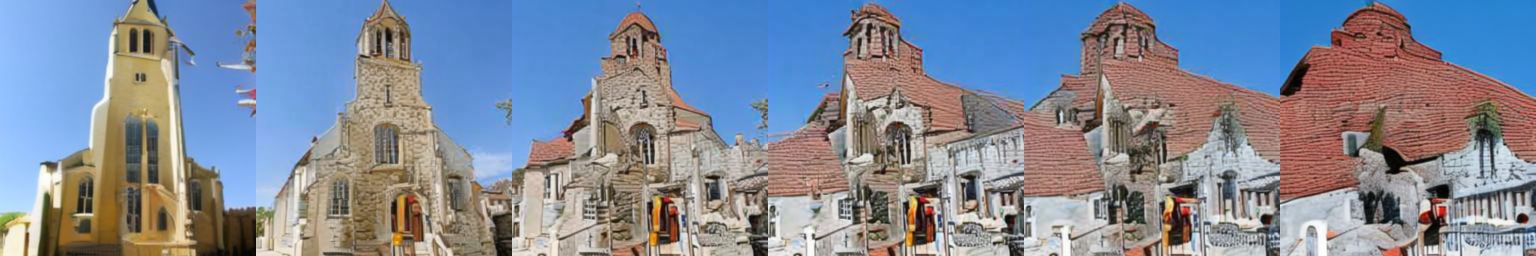} \\[1pt]
    \includegraphics[width=0.6\linewidth]{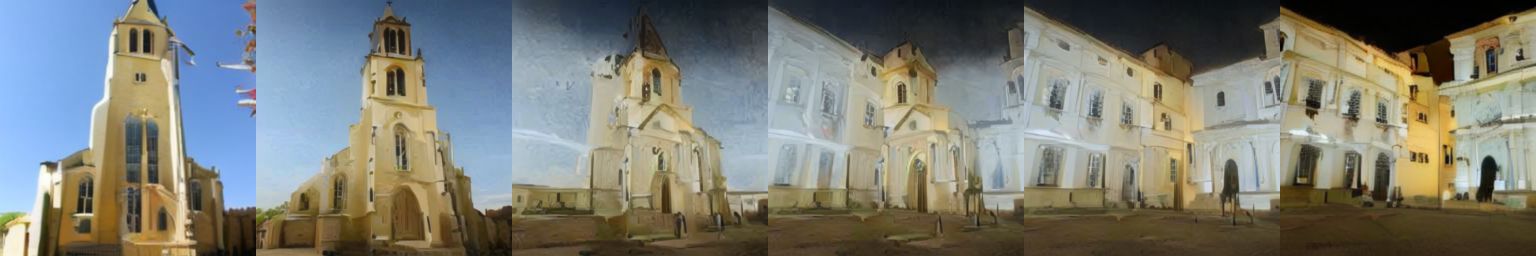}
    \vspace{-1mm}
    \caption{\small \textbf{Examples of low-quality trajectories from the LSUN-Church dataset.}}
    \label{fig:bad_samples}

    \vspace{4mm} 

    \begin{subfigure}[t]{0.42\linewidth}
        \centering
        \includegraphics[width=\linewidth]{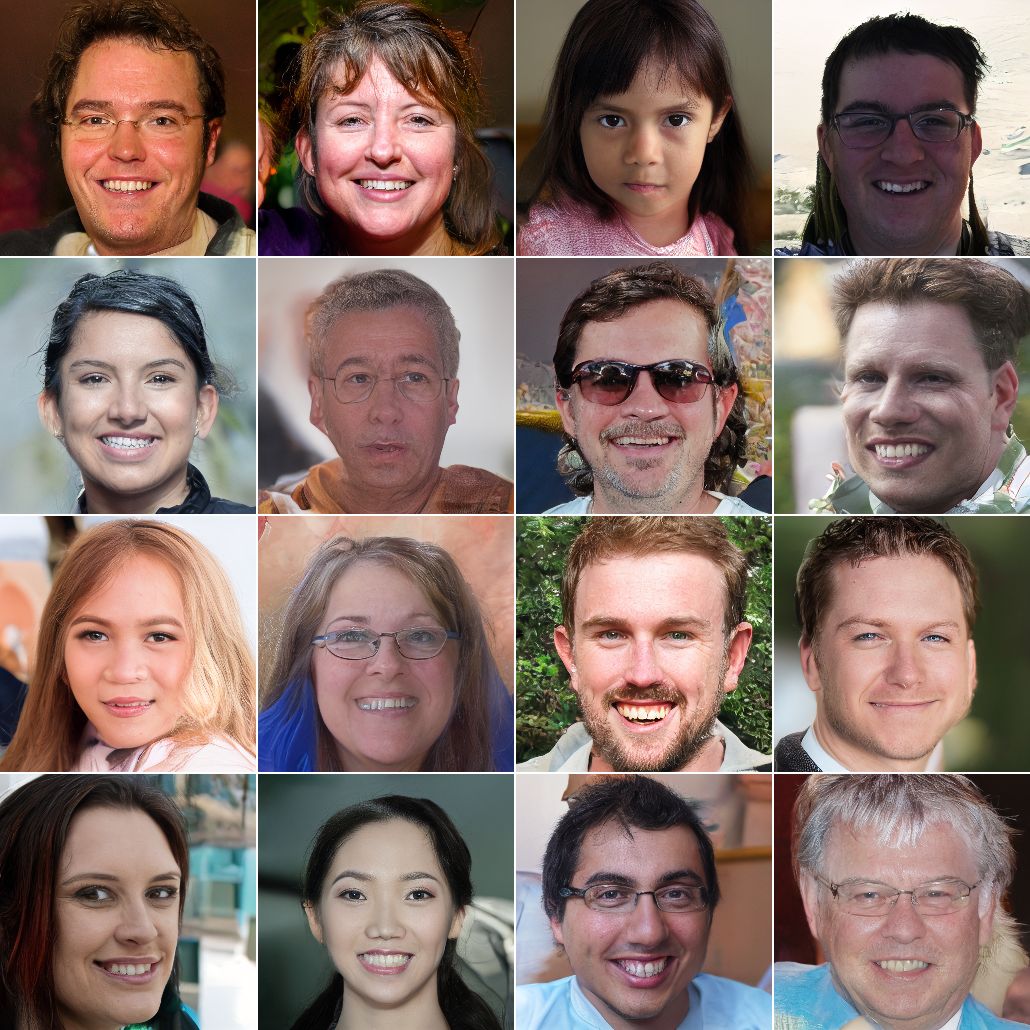}
        \caption{Baseline (FID = 5.032)}
        \label{fig:baseline}
    \end{subfigure}
    \hspace{0.03\linewidth}
    \begin{subfigure}[t]{0.42\linewidth}
        \centering
        \includegraphics[width=\linewidth]{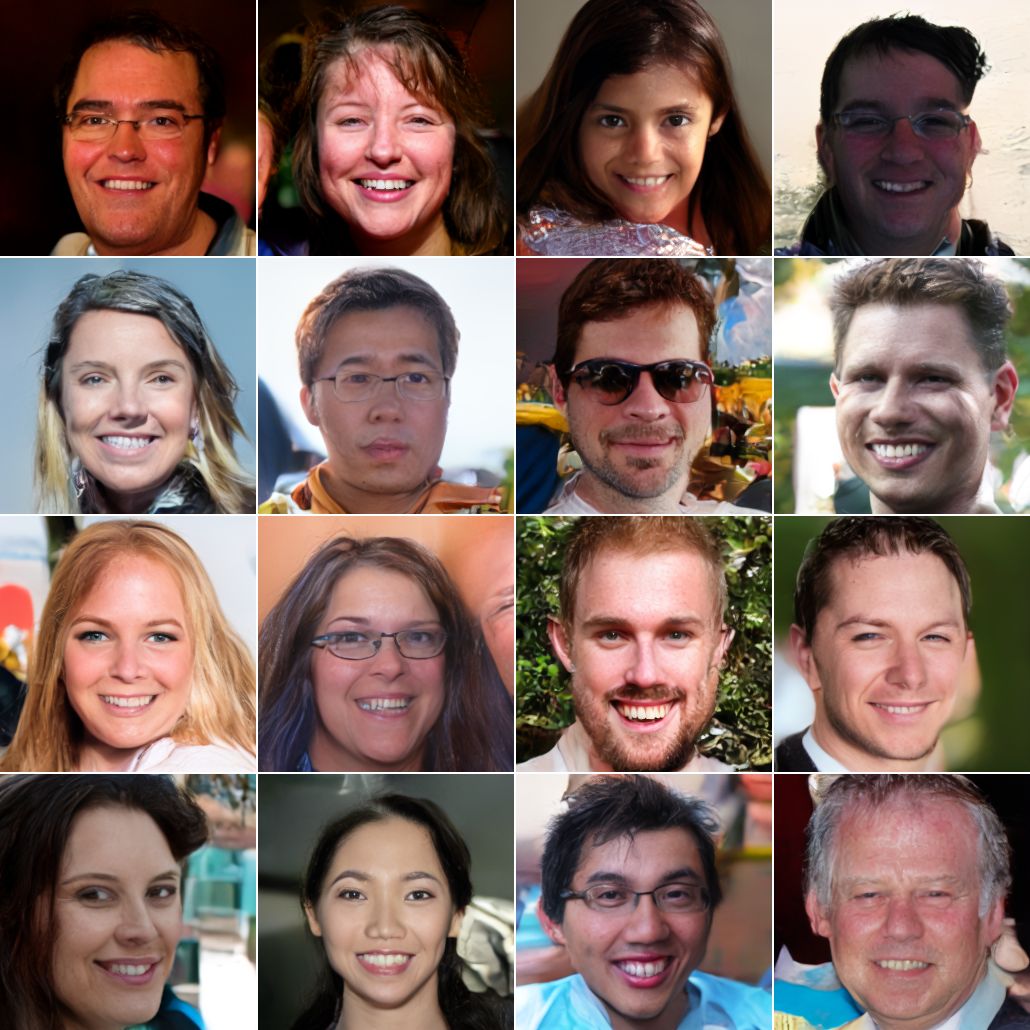}
        \caption{Ours (FID = 25.407)}
        \label{fig:ours}
    \end{subfigure}
    \vspace{-1mm}
    \caption{\small \textbf{Image quality comparison with the baseline.} Despite achieving a higher FID than the baseline (Lower FID is better), our model generates samples that remain recognizable and visually realistic.}
    \label{fig:fid_comparison}
\end{figure*}

The performance of our learned model depends on the quality of the morphing trajectory dataset. In this work, we employ DiffMorpher~\cite{zhang2024diffmorpher} to generate these trajectories because it produces smooth transitions and generalizes across diverse concepts. However, it does not guarantee that every intermediate state lies on the image manifold. This limitation is particularly evident for datasets with substantial structural and appearance variation, such as LSUN-Churches, where some trajectories contain distorted church structures or unrealistic intermediate images as shown in Fig.~\ref{fig:bad_samples}. Moreover, the trajectory quality is also limited by the quality of the pretrained LDM-generated endpoint images, whose artifacts can propagate through the trajectories. However, the morphing algorithm is used only as a preprocessing step, making our framework compatible with future improvements or alternative morphing methods.

Additionally, our prototype selection strategy relies on CLIP to identify a representative image. Future work could explore alternative approaches for obtaining prototypes that better reside on the data manifold, such as average image~\cite{Thawatdamrongkit_2026_CVPR, Feng_2025_CVPR}, dataset distillation~\cite{chan2025mgd3, Su_2024_CVPR, gu2024efficient}, and selecting representatives from the learned semantic latent space of generative models~\cite{preechakul2022diffae, karras2019stylegan, karras2020stylegan2, hudson2023soda}.



\vspace{-1.2em}
\subsubsection{Generated image quality.}
While our model generates smooth and coherent trajectories, our final generated images achieve a higher FID (25.407) than the baseline (5.032), which is initialized from an LDM pretrained on FFHQ and fine-tuned on our trajectory dataset. As shown in Fig.~\ref{fig:fid_comparison}, our model tends to produce images that contain fewer high-frequency details than those generated by the baseline, contributing to the degradation in FID. Despite this, the generated samples remain visually realistic and qualitatively similar to those of the baseline.

\vspace{-1em}
\subsubsection{Computation.}
Due to limited computational resources, we mainly conduct our experiments on the FFHQ dataset. The hyperparameters for the other datasets have not been extensively tuned, and more careful optimization may further improve their performance.

%% file: sec/7_conclusion.tex
\section{Conclusion}
We present Speed-Variation Distillation, a novel generative modeling idea. By leveraging variation-reduction trajectories obtained through image morphing together with our distillation framework, our method learns to transform a single shared prototype into high-quality and diverse images through a sequence of realistic intermediate states. This capability introduces a novel speed-diversity trade-off, enabling faster image generation at the cost of sample diversity.